\documentclass[10pt,twocolumn,letterpaper]{article}
\usepackage[pagenumbers]{cvpr}

\usepackage{times}
\usepackage{epsfig}
\usepackage{graphicx}
\usepackage{amsmath,amssymb,amsfonts,amsthm}
\usepackage{bm}
\usepackage{booktabs}
\usepackage{multirow}
\usepackage{xcolor}
\usepackage{colortbl}
\usepackage{array}
\usepackage{tabularx}
\usepackage{enumitem}
\usepackage{xspace}
\usepackage[ruled,vlined]{algorithm2e}
\SetKwInput{KwInput}{Input}
\SetKwInput{KwOutput}{Output}

\newtheorem{proposition}{Proposition}
\newtheorem{theorem}{Theorem}
\newtheorem{lemma}{Lemma}

\newtheorem{definition}{Definition}

\definecolor{oursbg}{RGB}{253,240,242}

\newcommand{\method}{\textsc{Tessal}\xspace}
\newcommand{\R}{\mathbb{R}}

\newcommand{\E}{\mathbb{E}}
\newcommand{\Var}{\mathrm{Var}}
\newcommand{\norm}[1]{\left\|#1\right\|}
\newcommand{\ip}[2]{\left\langle #1, #2 \right\rangle}
\newcommand{\tr}{\mathrm{tr}}

\newcommand{\vecop}{\mathrm{vec}}
\newcommand{\RFFT}{\mathrm{RFFT}}
\newcommand{\IRFFT}{\mathrm{IRFFT}}
\newcommand{\LMSE}{\mathcal{L}_{\mathrm{MSE}}}
\newcommand{\Ltilde}{\widetilde{\mathcal{L}}}
\newcommand{\Lvp}{\widetilde{\mathcal{L}}_{\mathrm{VP}}}
\newcommand{\Mten}{\mathcal{M}}
\newcommand{\Vten}{\mathcal{V}}
\newcommand{\Msk}{\widetilde{M}}
\newcommand{\Vsk}{\widetilde{V}}
\newcommand{\Hmap}{\mathcal{H}}

\makeatletter
\renewcommand{\paragraph}{\@startsection{paragraph}{4}{\z@}{0.7ex plus 0.2ex minus 0.1ex}{-0.8em}{\normalsize\bfseries}}
\makeatother

\definecolor{cvprblue}{rgb}{0.21,0.49,0.74}
\usepackage[pagebackref,breaklinks,colorlinks,allcolors=cvprblue]{hyperref}

\def\confName{CVPR}
\def\confYear{2026}
\def\paperID{0000}

\title{\texorpdfstring{\method{}}{TESSAL}: Chain Tensor-Sketched Deep Analytic Continual Learning}

\author{
Owen Whitaker\textsuperscript{1} \quad
Meilin Chen\textsuperscript{1} \quad
Arjun Subramanian\textsuperscript{2} \quad
Clara Lindholm\textsuperscript{1} \quad
Henrik Solberg\textsuperscript{2} \\
\textsuperscript{1}University of Bristol \qquad \textsuperscript{2}ETH Zurich
}

\begin{document}
\maketitle

\begin{abstract}
Analytic class-incremental learning (ACIL) achieves exact equivalence to full-replay training---\emph{absolute memorization}---by maintaining additive second-order sufficient statistics over frozen pre-trained features. Yet this guarantee has remained restricted to single-layer linear heads: extending exact sufficient statistics to depth-$L$ nonlinear networks causes the moment tensor to scale as $O((d{+}1)^{2L})$, reaching hundreds of terabytes even at $L{=}2$. We show that deep analytic continual learning can be realized in strictly polynomial space and wall-clock seconds without storing exemplars or optimizing synthetic surrogates. Our key insight is to pair \emph{input-coupled multilinear} heads---decoupled-gate MLPs, GaLU, and $\Pi$-Nets---with a recursive left-to-right \emph{Chain TensorSketch} that compresses the $L$-fold outer product into additive $O(m^2)$ sufficient statistics in a single streaming pass, independent of both sample count $N$ and depth $L$. On the model side, we prove that contracting the exponentially large parameter tensor through the exact same sketch requires \emph{zero parameter-side FFTs}, collapsing to $L{-}1$ sparse shift-and-hash steps. Theoretically, Chain TensorSketch reduces depth variance from flat sketching's $O(3^L/m)$ blow-up to linear $O(L/m)$, induces a non-negative implicit functional tensor-norm regularizer, and admits exact closed-form final-layer elimination via Variable Projection with inverse-free Envelope gradients. Across seven ViT-B/16 benchmarks and three million-scale streams past the $N^*{\approx}130\text{k}$ storage crossover (ImageNet-1K, Places365, iNat21), \method{} outperforms shallow analytic heads by $+1.4$--$3.0$ pts at matched $0.40\,\text{GB}$ state while matching or surpassing $10\times$--$21\times$ larger full replay in under $30$ seconds.
\end{abstract}

\section{Introduction}
\label{sec:intro}

Class-incremental learning (CIL) requires a vision model to absorb a sequence of disjoint class sets without access to task identities at inference and under strict cross-task memory limits~\cite{zhou2024ptmsurvey,hou2026transition}. With the rise of strong pre-trained models (PTMs), \emph{analytic continual learning} has emerged as a compelling exemplar-free paradigm~\cite{zhuang2022acil,mcdonnell2023ranpac,zhuang2024dsal,peng2025loranpac,zhao2026vila,zou2026flycl}. By freezing a pre-trained encoder and fitting a ridge-regression classifier over expanded features, analytic learners summarize the entire task stream into two additive matrices---the feature autocorrelation $M = \E[xx^\top]$ and cross-correlation $V = \E[yx^\top]$---whose $O(d^2)$ storage is constant in dataset size $N$ and yields a closed-form solution identical to joint full-replay training.

\begin{figure*}[t]
    \centering
    \includegraphics[width=0.95\textwidth]{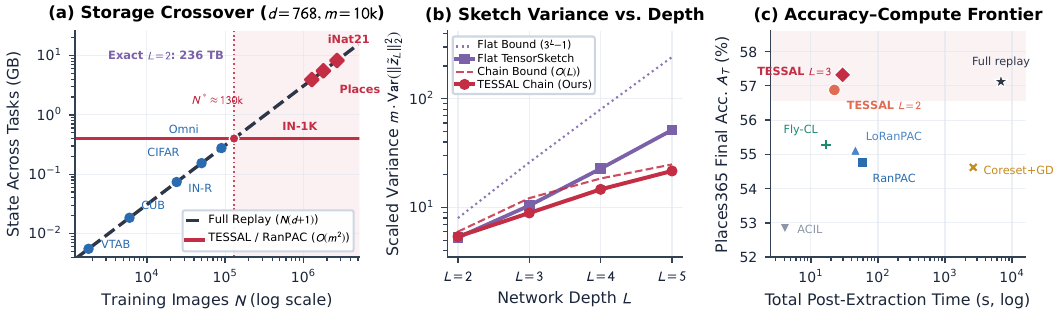}
    \caption{\textbf{Why deep analytic continual learning needs Chain TensorSketch.} \textbf{(a)} Honest cross-task state accounting at ViT-B/16 dimension $d{=}768$ and sketch width $m{=}10\text{k}$ ($0.40\,\text{GB}$). Full replay crosses the constant cache at $N^*{\approx}130\text{k}$, above all seven standard PTM-CIL benchmarks and $10\times$--$21\times$ below ImageNet-1K, Places365, and iNat21; exact $L{=}2$ deep moments sit off-chart at $236\,\text{TB}$. \textbf{(b)} Recursive Chain TensorSketch cuts norm-estimation variance from flat TensorSketch's exponential $O(3^L/m)$ blow-up to linear $O(L/m)$. \textbf{(c)} On Places365 ($0.40\,\text{GB}$), \method{} dominates shallow analytic methods, coresets, and full-batch GD on the accuracy--compute frontier.}
    \label{fig:motivation}
\end{figure*}

Despite rapid progress, every analytic CIL method remains bound to a fundamental expressivity ceiling: exact memorization holds only for a \emph{shallow ($L{=}1$) linear head} over fixed features. Even when features are lifted through wide random projections~\cite{mcdonnell2023ranpac,peng2025loranpac,zou2026flycl}, the classifier is restricted to a fixed kernel regression whose underfitting worsens under severe domain shift and fine-grained classes~\cite{zhuang2024dsal,zhao2026vila}. Can we extend the exact sufficient-statistic property of analytic CL to \emph{deep multi-layer neural networks} while preserving constant-in-$N$ storage and fast closed-form solving?

A natural structural bridge is provided by \emph{input-coupled multilinear} architectures---including Gated Linear Units (GaLU)~\cite{fiat2019galu}, $\Pi$-Nets~\cite{chrysos2020pinets}, Multilinear Operator Networks~\cite{cheng2024monet}, and our depth-$L$ Decoupled-Gate MLP (DG-MLP). Because every gating map $\phi_\ell(x){=}\sigma(F_{\ell-1}x)$ depends only on the frozen input $x$ rather than preceding hidden states, the network output separates cleanly into a Frobenius inner product $f_\theta(x)_c = \ip{G(\theta)_c}{P(x)}$ between a learnable parameter tensor $G(\theta)_c$ and a data tensor $P(x) = \bigotimes_{\ell=1}^L \phi_\ell(x)$. Consequently, the \emph{entire} MSE loss landscape and its exact gradients are determined by two additive moment tensors $(\Mten_T, \Vten_T)$ (Theorem~\ref{thm:exact-sufficiency}).

However, this exact formulation hits an immediate exponential wall: $\Mten_T$ has $O((d{+}1)^{2L})$ entries---roughly $236\,\text{TB}$ at $L{=}2$ for ViT-B/16 ($d{=}768, k_2{=}10\text{k}$) and $820\,\text{PB}$ at $L{=}3$ (Figure~\ref{fig:motivation}a). Worse, naive compression via flat $L$-mode TensorSketch~\cite{pham2025tensorsketch} suffers exponential $O(3^L/m)$ variance blow-up with depth (Figure~\ref{fig:motivation}b), while iterative surrogate distillation or full-batch GD sacrifices the single-pass efficiency and closed-form speed that motivate analytic CL in the first place.

We resolve all three bottlenecks with \textbf{\method{}} (\textbf{Te}nsor-\textbf{S}ketched \textbf{S}ufficient-Statistic \textbf{A}nalytic \textbf{L}earning). First, inspired by oblivious recursive polynomial sketching~\cite{ahle2020oblivious}, \method{} folds the $L$ input-coupled maps left-to-right through a \emph{Chain TensorSketch} $z_\ell(x) = S_\ell(\phi_\ell(x) \otimes z_{\ell-1}(x)) \in \R^m$, accumulating additive sufficient statistics $(\Msk_T, \Vsk_T)$ of size $O(m^2{+}mC)$ in a \emph{single streaming pass} with zero per-task optimization. Second, we prove an algebraic surprise on the parameter side: because each layer's gate basis $e_k$ hashes to a $1$-sparse vector, contracting the exponentially large tensor $G(\theta)$ through the exact same Chain TensorSketch requires \emph{zero parameter-side FFTs}, reducing to $L{-}1$ sparse shift-and-hash operations (Proposition~\ref{prop:shift-hash}). Third, we establish that unbiased sketching yields an \emph{implicit functional tensor-norm regularizer} $\mathcal{R}_m(\theta) \le \frac{ec_2(L-1)}{m}\sum_c \|G(\theta)_c\|_F^2 \E\|P(x)\|_2^2$ (Theorem~\ref{thm:variance-reg}), explaining both asymptotic exactness as $m{\to}\infty$ and generalization gains over unregularized replay at practical budgets. Finally, exploiting exact linearity in the output weights $W_L$, we introduce a \emph{Variable-Projected Analytic Solver} (VarPro-ACIL)~\cite{golub2003varpro} that eliminates $W_L^*$ in closed form in $O(\min(k_L,m)^3)$ time and updates $W_{1:L-1}$ via inverse-free Envelope gradients (Theorem~\ref{thm:varpro}), converging in $80$ steps even at $C{=}10{,}000$ classes (Figure~\ref{fig:motivation}c).

Our contributions are fourfold:
\begin{itemize}[leftmargin=1.2em,itemsep=1pt,topsep=1.5pt]
    \item \textbf{Single-Pass Deep Analytic Memory:} Chain TensorSketch sufficient statistics cut storage from $O((d{+}1)^{2L})$ to $O(m^2{+}mC)$ (independent of $N,L$) with exact additivity and zero per-task optimization.
    \item \textbf{Zero-FFT Parameter Contraction:} An exact sparse shift-and-hash operator $\Hmap_\ell$ contracts depth-$L$ parameter tensors into $\R^{C\times m}$ without FFTs or tensor materialization.
    \item \textbf{Theory \& VarPro Elimination:} We prove $O(L/m)$ variance (vs.\ flat $O(3^L/m)$), identify the implicit functional tensor-norm regularizer $\mathcal{R}_m(\theta)$, and derive closed-form $W_L^*$ elimination with inverse-free Envelope gradients.
    \item \textbf{State-of-the-Art Across Scales:} Across seven ViT-B/16 benchmarks and three million-scale streams past $N^*{\approx}130\text{k}$, \method{} sets new SOTA at matched $0.40\,\text{GB}$ state while running $100\times$ faster than iterative deep heads.
\end{itemize}

\section{Related Work}
\label{sec:related}

\paragraph{Pre-trained model class-incremental learning.}
PTM-based CIL has shifted the field from full-backbone rehearsal toward lightweight adaptation over frozen Vision Transformers~\cite{zhou2024ptmsurvey,hou2026transition}. Prompting methods such as CODA-Prompt~\cite{smith2023codaprompt} isolate plasticity in learnable attention-weighted prompts, while adapter and low-rank pipelines---including EASE~\cite{zhou2024ease}, SD-LoRA~\cite{wu2025sdlora}, CL-LoRA~\cite{he2025cllora}, FlyLoRA~\cite{zou2026flylora}, TUNA~\cite{wang2025tuna}, and joint-loss LoRA merging~\cite{gu2026merging}---expand or merge parameter-efficient subspaces across sessions. Complementary efforts align sequential classifiers~\cite{zhang2024slcapp} or tap frozen discriminative geometry directly~\cite{wang2026tap}. Despite strong results, iterative gradient updates over prompts or adapters remain sensitive to task order and lack closed-form replay equivalence.

\paragraph{Analytic continual learning.}
Starting from ACIL~\cite{zhuang2022acil}, analytic CIL replaces backpropagation with recursive ridge regression over frozen features, offering deterministic updates and exact full-replay equivalence. Subsequent work broadened this paradigm via dual-stream residual learning (DS-AL)~\cite{zhuang2024dsal}, generalized and forward-only online updates (GACL, F-OAL)~\cite{zhuang2024gacl,zhuang2024foal}, and representation enhancement (REAL)~\cite{he2024real}. On pre-trained ViTs, RanPAC~\cite{mcdonnell2023ranpac} and LoRanPAC~\cite{peng2025loranpac} lift features through wide or low-rank random projections, while Fly-CL~\cite{zou2026flycl,zou2025flycircuit} and GRP~\cite{li2026grp} accelerate decorrelation via sparse or guided expansions. Recent 2025--2026 advances further introduced latent calibration (CIFNet)~\cite{dopico2025cifnet}, analytic contrastive projection (AnaCP)~\cite{momeni2025anacp}, vision-language calibration (VILA)~\cite{zhao2026vila}, low-rank RGDA with Hopfield drift compensation~\cite{rao2026scalable}, spectral rectification for long-tailed streams (GSR)~\cite{tran2026spectral}, equinorm remedies for ETF collapse~\cite{sezer2026impossibility}, unlabeled posterior analytics~\cite{tang2026posterior}, non-kernel diagonal sparsity~\cite{karpel2026beyond}, and graph extensions~\cite{song2026drift,chen2026magic}. Crucially, all of these analytic methods still solve a \emph{single-layer} linear or quadratic discriminant head; \method{} removes this depth-$1$ barrier while keeping single-pass additive statistics and closed-form solving.

\paragraph{Multilinear networks and recursive tensor sketching.}
Decoupled-gate networks (GaLU)~\cite{fiat2019galu} separate neuron gating from linear weights, while $\Pi$-Nets~\cite{chrysos2020pinets} and Multilinear Operator Networks~\cite{cheng2024monet} express deep representations as high-degree polynomials with rich recent theory on identifiability~\cite{usevich2025identifiability}, robustness~\cite{alexandr2026robustness}, minimal filling~\cite{dao2026minimal}, and algebraic expressivity~\cite{akbari2026algebraic}. To compress outer-product feature maps without exponential blow-up, TensorSketch~\cite{pham2025tensorsketch} applies FFT-based circular convolution of CountSketches, and recursive tree/chain constructions~\cite{ahle2020oblivious,khan2026trace} eliminate exponential-in-degree variance, powering fast polynomial attention~\cite{kacham2024polysketchformer}. \method{} is the first framework to unify multilinear neural architectures, recursive Chain TensorSketch, zero-FFT parameter contraction, and Variable Projection~\cite{golub2003varpro} for deep analytic continual learning.

\section{Preliminaries and Multilinear Sufficiency}
\label{sec:prelim}

\paragraph{Class-incremental protocol and notation.}
A learner observes a stream of $T$ tasks $\mathcal{D}_1,\dots,\mathcal{D}_T$ with disjoint label sets $\mathcal{Y}_1,\dots,\mathcal{Y}_T$ and total classes $C = \sum_{t=1}^T |\mathcal{Y}_t|$. A pre-trained encoder $\psi:\mathcal{I}\to\R^d$ maps each image $I$ to a frozen $\ell_2$-normalized feature $x=\psi(I)\in\R^d$ paired with a one-hot target $y\in\R^C$. Throughout, $\odot$ denotes the Hadamard product, $\otimes$ the tensor outer product, $\ip{\cdot}{\cdot}$ the Frobenius inner product, and $P^{\otimes 2}:=P\otimes P$. For clarity of exposition we assume equal task sizes $N/T$ (generalizing trivially via sample counts $N_t$).

\paragraph{Shallow analytic CL ($L{=}1$).}
In ACIL~\cite{zhuang2022acil} and RanPAC~\cite{mcdonnell2023ranpac}, a frozen expansion $\phi:\R^d\to\R^M$ feeds a single linear layer $f_W(x)=W\phi(x)$. Minimizing MSE risk $\LMSE(W)=\E[\norm{W\phi(x)-y}_2^2]+\gamma\norm{W}_F^2$ yields $W^* = V(M+\gamma I)^{-1}$, where $M=\E[\phi(x)\phi(x)^\top]\in\R^{M\times M}$ and $V=\E[y\phi(x)^\top]\in\R^{C\times M}$ accumulate additively across tasks. Because $(M,V)$ determine $\LMSE(W)$ identically everywhere, the streaming learner recovers the exact joint-training minimizer without storing raw data.

\paragraph{Input-coupled multilinear networks (ICMNs).}
To move beyond $L{=}1$ while preserving sufficient-statistic separability, we restrict hidden nonlinearities to depend on the \emph{unchanging input $x$} rather than prior learnable activations.

\begin{definition}[Degree-$L$ Input-Coupled Multilinear Network]
\label{def:icmn}
Fix depth $L{\ge}2$, widths $k_1,\dots,k_L{\in}\mathbb{N}_{\ge 2}$, and parameter-free anchor maps $\Phi=(\phi_\ell:\R^d{\to}\R^{k_\ell})_{\ell=1}^L$. A network $f_\theta:\R^d{\to}\R^C$ smooth in $\theta{\in}\Theta$ is degree-$L$ input-coupled multilinear if there exists a unique family of smooth coefficient tensors $\{g(\theta,A)\}_{A\subseteq [L]}$ with $g(\theta,A)_c\in\bigotimes_{\ell\in A}\R^{k_\ell}$ such that
\begin{equation}
\label{eq:icmn-expansion}
f_\theta(x)_c = \sum_{A\subseteq [L]} \ip{g(\theta,A)_c}{\bigotimes_{\ell\in A}\phi_\ell(x)}, \qquad \forall c\in[C],
\end{equation}
and the top-order block satisfies $\mathrm{span}\{g(\theta,[L])_c:\theta\in\Theta\} = \bigotimes_{\ell=1}^L \R^{k_\ell}$.
\end{definition}

Our primary instantiation is the \textbf{Decoupled-Gate MLP (DG-MLP)}: fix $k_1{=}d$, frozen random gate matrices $F_1,\dots,F_{L-1}$ ($F_{\ell-1}\in\R^{k_\ell\times d}$), $\phi_1(x)=x$, and $\phi_\ell(x)=\sigma(F_{\ell-1}x)$ for $\ell{\ge}2$, with learnable weights $W_\ell\in\R^{k_{\ell+1}\times k_\ell}$ ($k_{L+1}{=}C$):
\begin{align}
\label{eq:dgmlp}
h_1(x) &= W_1\phi_1(x), \quad f_\theta(x)=h_L(x), \nonumber \\
h_\ell(x) &= W_\ell\big(\phi_\ell(x)\odot h_{\ell-1}(x)\big) \quad (\ell=2,\dots,L).
\end{align}
DG-MLP has a single active subset $\mathcal{A}=\{[L]\}$: its parameter tensor $G(\theta)_c\in\R^{k_L\times\dots\times k_1}$ has entries
\begin{equation}
\label{eq:dgmlp-tensor}
G(\theta)_{c,i_L,\dots,i_1} = (W_L)_{c,i_L}(W_{L-1})_{i_L,i_{L-1}}\cdots(W_1)_{i_2,i_1},
\end{equation}
and $f_\theta(x)_c = \ip{G(\theta)_c}{P(x)}$ where $P(x):=\bigotimes_{\ell=1}^L \phi_\ell(x)$. GaLU~\cite{fiat2019galu} is the special case $L{=}2,\sigma(u)=\mathbf{1}[u{>}0]$, and $\Pi$-Net CCP~\cite{chrysos2020pinets} also satisfies Definition~\ref{def:icmn} (proved in Appendix~\ref{app:proofs}).

\begin{theorem}[Deep Multilinear Sufficiency and Exponential Wall]
\label{thm:exact-sufficiency}
Let $f_\theta$ be a degree-$L$ ICMN with $G(\theta)_c := \bigoplus_{A\in\mathcal{A}} \vecop(g(\theta,A)_c)$ and $P(x) := \bigoplus_{A\in\mathcal{A}} \vecop(\bigotimes_{\ell\in A}\phi_\ell(x))$. Define additive moment tensors $\Mten_T := \frac{1}{T}\sum_{t=1}^T \E_{\mathcal{D}_t}[P(x)^{\otimes 2}]$ and $\Vten_{T,c} := \frac{1}{T}\sum_{t=1}^T \E_{\mathcal{D}_t}[y_c P(x)]$. Then for every $\theta\in\Theta$,
\begin{equation}
\label{eq:exact-loss}
\LMSE(\theta) = \sum_{c=1}^C \Big(\ip{G(\theta)_c^{\otimes 2}}{\Mten_T} - 2\ip{G(\theta)_c}{\Vten_{T,c}}\Big) + \E\norm{y}_2^2.
\end{equation}
Hence full-batch optimization on $(\Mten_T,\Vten_T)$ reproduces full-replay training identically. However, at uniform width $k_\ell{=}d$, storing $\Mten_T$ requires $\Theta((d{+}1)^{2L})$ floats (or $d^{2L}$ for DG-MLP).
\end{theorem}

Theorem~\ref{thm:exact-sufficiency} clarifies the exact dilemma: multilinear gating breaks the depth-$1$ barrier of ACIL, yet exact moment tensors are physically unstoreable ($236\,\text{TB}$ at $L{=}2$ for $d{=}768, k_2{=}10\text{k}$). Next, we compress $(\Mten_T,\Vten_T)$ into $O(m^2)$ space without sacrificing single-pass additivity or analytic solving.

\section{\texorpdfstring{\method{}}{TESSAL}}
\label{sec:method}

\begin{figure*}[t]
    \centering
    \includegraphics[width=0.94\textwidth]{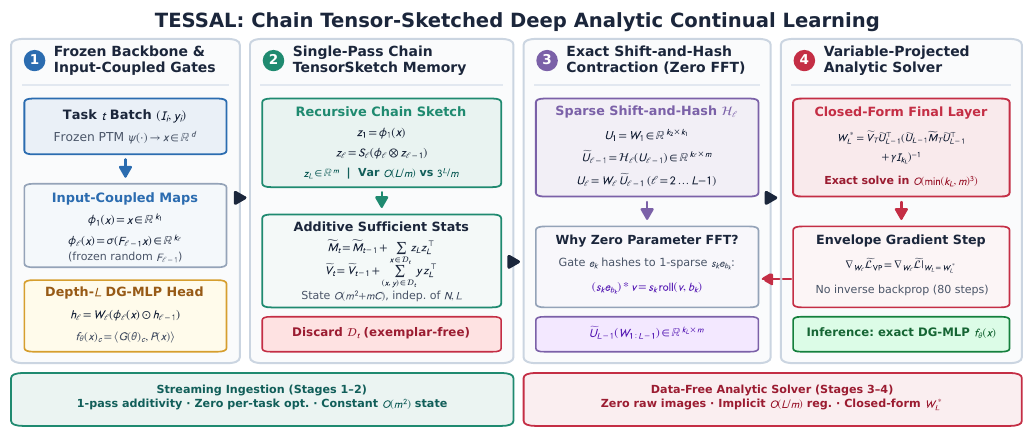}
    \caption{\textbf{Overview of \method{}.} \textbf{Stages 1--2 (Streaming Ingestion):} Frozen backbone features $x{=}\psi(I)$ generate input-coupled maps $\phi_\ell(x)$ that are folded left-to-right via a recursive degree-$2$ Chain TensorSketch into $z_L(x){\in}\R^m$, accumulating exact additive statistics $(\Msk_t,\Vsk_t)$ of depth-independent size $O(m^2{+}mC)$ before discarding $\mathcal{D}_t$. \textbf{Stages 3--4 (Data-Free Analytic Solver):} Weights $W_{1:L-1}$ are contracted through the exact same sketch using zero-FFT sparse shift-and-hash steps $\Hmap_\ell$; the final layer $W_L^*$ is eliminated in closed form via Variable Projection while $W_{1:L-1}$ updates via inverse-free Envelope gradients.}
    \label{fig:architecture}
\end{figure*}

Figure~\ref{fig:architecture} summarizes \method{}, which separates continual learning into a single-pass \emph{Streaming Ingestion Path} (Stages 1--2) and a data-free \emph{Analytic Solver Path} (Stages 3--4).

\subsection{Single-Pass Chain TensorSketch Memory}
\label{sec:chain-sketch}

A flat $L$-mode TensorSketch~\cite{pham2025tensorsketch} hashes all $L$ anchor maps simultaneously and convolves them in one FFT step, incurring variance that grows exponentially as $O(3^L/m)$ with depth $L$. Instead, motivated by recursive polynomial sketching~\cite{ahle2020oblivious}, we compose $L{-}1$ independent degree-$2$ TensorSketches along the network chain.

For each layer $\ell \in \{2,\dots,L\}$, draw two independent 2-wise hash maps $h_{\ell,\phi}:[k_\ell]\to[m]$, $h_{\ell,z}:[m_{\ell-1}]\to[m]$ (with $m_1{=}k_1$ and $m_\ell{=}m$ for $\ell{\ge}2$) and independent 4-wise Rademacher sign maps $s_{\ell,\phi}:[k_\ell]\to\{\pm 1\}$, $s_{\ell,z}:[m_{\ell-1}]\to\{\pm 1\}$. Let $C_{\ell,\phi}\in\R^{m\times k_\ell}$ and $C_{\ell,z}\in\R^{m\times m_{\ell-1}}$ denote the corresponding CountSketch operators. Starting from $z_1(x) = \phi_1(x)\in\R^{k_1}$, we recursively fold each anchor map into $\R^m$:
\begin{align}
\label{eq:chain-data}
z_\ell(x) &:= S_\ell\big(\phi_\ell(x) \otimes z_{\ell-1}(x)\big) \nonumber \\
&= \IRFFT\!\Big(\RFFT(C_{\ell,\phi}\phi_\ell(x)) \odot \RFFT(C_{\ell,z}z_{\ell-1}(x))\Big).
\end{align}
The terminal embedding $z_L(x)\in\R^m$ costs only $O(\sum_{\ell=2}^L (k_\ell + m\log m))$ time per image and never forms any outer product. When task $\mathcal{D}_t$ arrives, \method{} updates two matrices in a single streaming pass:
\begin{align}
\label{eq:additive-stats}
\Msk_t &= \Msk_{t-1} + \sum_{x\in\mathcal{D}_t} z_L(x) z_L(x)^\top \in \R^{m\times m}, \nonumber \\
\Vsk_t &= \Vsk_{t-1} + \sum_{(x,y)\in\mathcal{D}_t} y\, z_L(x)^\top \in \R^{C\times m},
\end{align}
normalizes by the total sample count $N_t = \sum_{s=1}^t |\mathcal{D}_s|$ when evaluating risk, and immediately discards $\mathcal{D}_t$. Storage is strictly $O(m^2{+}mC)$ floats---identical to a single-layer RanPAC cache of width $m$, completely independent of both sample count $N$ and network depth $L$, and requiring \emph{zero} per-task optimization.

\subsection{Exact Shift-and-Hash Parameter Contraction}
\label{sec:shift-hash}

Evaluating the sketched risk directly from $(\Msk_T, \Vsk_T)$ requires mapping the learnable weights $\theta = (W_1,\dots,W_L)$ into the \emph{exact same} Chain TensorSketch space $\R^{C\times m}$ without ever forming the exponential tensor $G(\theta)$. At first sight, sketching $G(\theta)$ appears to require $L{-}1$ batched FFTs over all rows of every weight matrix. Remarkably, the multilinear structure of Eq.~\eqref{eq:dgmlp-tensor} eliminates every parameter-side FFT.

Define the layer-$\ell$ \textbf{shift-and-hash operator} $\Hmap_\ell:\R^{k_\ell \times m_{\ell-1}} \to \R^{k_\ell \times m}$ acting on any matrix $U \in \R^{k_\ell \times m_{\ell-1}}$ as follows:
\begin{enumerate}[leftmargin=1.35em,itemsep=1pt,topsep=1.5pt]
    \item Hash columns via $C_{\ell,z}$: $\bar{U} := U C_{\ell,z}^\top \in \R^{k_\ell \times m}$ (a sparse scatter-add over $m_{\ell-1}$ columns).
    \item Circularly shift each row $k\in[k_\ell]$ by $h_{\ell,\phi}(k)$ with sign $s_{\ell,\phi}(k)$:
    \begin{equation}
    \label{eq:shift-hash}
    \big[\Hmap_\ell(U)\big]_{k,\, r} := s_{\ell,\phi}(k)\, \bar{U}_{k,\, (r - h_{\ell,\phi}(k)) \bmod m}.
    \end{equation}
\end{enumerate}
For $\ell{=}2$, steps 1--2 fuse into a single sparse scatter directly from $U_1{=}W_1\in\R^{k_2\times k_1}$ into bucket $(h_{2,\phi}(k)+h_{2,z}(j))\bmod m$ at cost $O(k_2 k_1)$. Applying $\Hmap_\ell$ recursively across layers produces the exact contracted parameter map:
\begin{equation}
\label{eq:param-chain}
U_1 = W_1,\quad \widetilde{U}_{\ell-1} = \Hmap_\ell(U_{\ell-1}),\quad U_\ell = W_\ell \widetilde{U}_{\ell-1},
\end{equation}
for $\ell=2,\dots,L$, so that $\widetilde{G}(\theta) := U_L = W_L \widetilde{U}_{L-1}(W_{1:L-1}) \in \R^{C\times m}$.

\begin{proposition}[Zero-FFT Exact Parameter Contraction]
\label{prop:shift-hash}
Let $z_L(x)$ be computed by Eq.~\eqref{eq:chain-data} and $\widetilde{G}(\theta)=U_L(\theta)$ by Eqs.~\eqref{eq:shift-hash}--\eqref{eq:param-chain} using the same hashes $\{(h_{\ell,\phi},s_{\ell,\phi},h_{\ell,z},s_{\ell,z})\}_{\ell=2}^L$. Then:
\begin{enumerate}[label=(\roman*),leftmargin=1.4em,itemsep=1pt,topsep=1pt]
    \item \emph{Exact Chain-Sketch Equivalence:} For every $c{\in}[C]$, $\widetilde{G}(\theta)_{c,:} = (S_L \circ \cdots \circ S_2)(G(\theta)_c)$ identically, and $\E_{S_{2:L}}[\ip{\widetilde{G}(\theta)_{c,:}}{z_L(x)}] = f_\theta(x)_c$.
    \item \emph{Polynomial Complexity:} Computing $\widetilde{U}_{L-1}$ costs $O(k_2 k_1 + \sum_{\ell=3}^L k_\ell m + \sum_{\ell=2}^{L-1} k_{\ell+1} k_\ell m)$ arithmetic operations with zero FFTs and $O(m \max_\ell k_\ell)$ working memory.
\end{enumerate}
\end{proposition}

The intuition is immediate: in Eq.~\eqref{eq:dgmlp}, unit $k$ of layer $\ell$ multiplies $\phi_\ell(x)_k = \ip{e_k}{\phi_\ell(x)}$ against row $k$ of $U_{\ell-1}$. Under CountSketch, the standard basis vector $e_k$ maps to the $1$-sparse impulse $s_{\ell,\phi}(k) e_{h_{\ell,\phi}(k)}$, and circularly convolving any vector $v\in\R^m$ with a $1$-sparse impulse is simply a signed circular shift---requiring zero Fourier transforms.

\subsection{Variable-Projected Analytic Solver (VarPro-ACIL)}
\label{sec:varpro}

Given the normalized sketched moments $(\Msk_T, \Vsk_T)$ (divided by $N_T$), the sketched MSE risk with ridge penalty $\gamma > 0$ on the output layer $W_L$ is
\begin{align}
\label{eq:sketched-risk}
\Ltilde(W_{1:L-1}, W_L) &= \tr\!\Big(W_L \widetilde{U}_{L-1} \Msk_T \widetilde{U}_{L-1}^\top W_L^\top\Big) \nonumber \\
&\quad - 2\,\tr\!\Big(W_L \widetilde{U}_{L-1} \Vsk_T^\top\Big) + \gamma\norm{W_L}_F^2 + \mathrm{const},
\end{align}
where $\widetilde{U}_{L-1} = \widetilde{U}_{L-1}(W_{1:L-1}) \in \R^{k_L \times m}$ is purely real and independent of $W_L$. Crucially, Eq.~\eqref{eq:sketched-risk} is a \emph{separable nonlinear least-squares} problem~\cite{golub2003varpro}: for any hidden weights $W_{1:L-1}$, the optimal classifier head $W_L^*$ is given in exact closed form by
\begin{equation}
\label{eq:varpro-wl}
W_L^*(W_{1:L-1}) = \Vsk_T \widetilde{U}_{L-1}^\top \big(\widetilde{U}_{L-1} \Msk_T \widetilde{U}_{L-1}^\top + \gamma I_{k_L}\big)^{-1}.
\end{equation}
Notice that $\widetilde{U}_{L-1}\Msk_T\widetilde{U}_{L-1}^\top + \gamma I_{k_L}$ is a $k_L \times k_L$ symmetric positive-definite (SPD) matrix (e.g., $2048\times 2048$ when $k_L{=}2048 < m{=}10\text{k}$), solvable via Cholesky in under $2\,\text{ms}$ on a GPU. Whenever $k_L > m$, the push-through identity (Theorem~\ref{thm:varpro}) flips the inversion to $m\times m$, guaranteeing $O(\min(k_L,m)^3)$ cost.

Minimizing the reduced Variable-Projection objective $\Lvp(W_{1:L-1}) := \Ltilde(W_{1:L-1}, W_L^*(W_{1:L-1}))$ keeps $W_L$ at its exact global ridge optimum at every step, eliminating the notorious ill-conditioning of joint SGD on many-class outputs ($C{=}10{,}000$ on iNat21). Even better, by the Envelope Theorem (Theorem~\ref{thm:varpro}), we compute exact gradients $\nabla_{W_\ell}\Lvp$ by simply evaluating $\nabla_{W_\ell}\Ltilde(W_{1:L-1}, W_L)\big|_{W_L=\mathrm{sg}[W_L^*]}$ with a stop-gradient on $W_L^*$---never backpropagating through the Cholesky solve. Algorithm~\ref{alg:tessal} details the complete streaming and solver procedure; at test time, inference executes the exact DG-MLP forward pass Eq.~\eqref{eq:dgmlp} with zero sketching overhead.

\begin{algorithm}[t]
\scriptsize
\caption{\method{} incremental update and VarPro-ACIL solve}
\label{alg:tessal}
\KwInput{Frozen PTM $\psi$, frozen gates $\{F_{\ell-1}\}_{\ell=2}^L$, Chain hashes $\{C_{\ell,\phi},C_{\ell,z}\}_{\ell=2}^L$, stats $(\Msk_{t-1},\Vsk_{t-1},N_{t-1})$, task data $\mathcal{D}_t$, ridge $\gamma$, outer steps $K$}
\KwOutput{Updated stats $(\Msk_t,\Vsk_t,N_t)$ and deep head weights $\theta_t=(W_1,\dots,W_L)$}

\tcp{Stage 1--2: Single-pass streaming Chain TensorSketch}
Initialize $\Delta\Msk \leftarrow 0$, $\Delta\Vsk \leftarrow 0$ \;
\For{mini-batch $(I, Y) \subset \mathcal{D}_t$}{
    Extract $X \leftarrow \psi(I)$; compute $\phi_1{=}X$, $\phi_\ell{=}\sigma(F_{\ell-1}X)$ ($\ell{=}2{\dots}L$) \;
    Fold $Z_1{=}\phi_1$, $Z_\ell \leftarrow S_\ell(\phi_\ell \otimes Z_{\ell-1})$ via FFT Eq.~\eqref{eq:chain-data} \;
    $\Delta\Msk \leftarrow \Delta\Msk + Z_L^\top Z_L$, \quad $\Delta\Vsk \leftarrow \Delta\Vsk + Y^\top Z_L$ \;
}
$\Msk_t \leftarrow \Msk_{t-1} + \Delta\Msk$, \quad $\Vsk_t \leftarrow \Vsk_{t-1} + \Delta\Vsk$, \quad $N_t \leftarrow N_{t-1} + |\mathcal{D}_t|$ \;
Discard $\mathcal{D}_t$; set $\bar{M} \leftarrow \Msk_t/N_t$, $\bar{V} \leftarrow \Vsk_t/N_t$ \;

\tcp{Stage 3--4: Data-free VarPro-ACIL solve}
\For{$s = 1,\dots,K$}{
    Contract $W_{1:L-1} \mapsto \widetilde{U}_{L-1}$ via zero-FFT $\Hmap_\ell$ Eq.~\eqref{eq:param-chain} \;
    Solve $W_L^* \leftarrow \bar{V}\widetilde{U}_{L-1}^\top(\widetilde{U}_{L-1}\bar{M}\widetilde{U}_{L-1}^\top + \gamma I_{k_L})^{-1}$ (Cholesky) \;
    Compute Envelope grad $g_\ell \leftarrow \nabla_{W_\ell}\Ltilde(W_{1:L-1}, \mathrm{sg}[W_L^*])$ \;
    Update $W_{1:L-1} \leftarrow \mathrm{AdamW}(W_{1:L-1}, \{g_\ell\})$ \;
}
Compute final exact $W_L^*$ at converged $W_{1:L-1}$ \;
\Return $(\Msk_t,\Vsk_t,N_t)$, $\theta_t = (W_1,\dots,W_{L-1},W_L^*)$ \;
\end{algorithm}

\section{Theoretical Analysis}
\label{sec:theory}

\begin{table*}[t]
\centering
\footnotesize
\renewcommand{\arraystretch}{0.92}
\setlength{\tabcolsep}{3.6pt}
\caption{\textbf{Suite 1: Last-task accuracy $A_T$ (\%) across seven ViT-B/16-IN21K PTM-CIL benchmarks}, plus mean $A_T$ and mean average incremental accuracy $\bar{A}$ (\%) over all eight settings (per-benchmark $\bar{A}$ in Appendix Table~\ref{tab:app-full}). All analytic/sketched methods match $0.40\,\text{GB}$ state ($M{=}m{=}10\text{k}$), except the $2\times$-width control ($1.60\,\text{GB}$). \textbf{Bold}: best; \underline{underline}: second-best.}
\label{tab:main}
\begin{tabular*}{\textwidth}{@{\extracolsep{\fill}}lcccccccccc@{}}
\toprule
\multirow{2}{*}{Method} & \multicolumn{2}{c}{CIFAR-100} & CUB & IN-R & IN-A & ObjectNet & Omni & VTAB & \multirow{2}{*}{Avg.\ $A_T$} & \multirow{2}{*}{Avg.\ $\bar{A}$} \\
\cmidrule(lr){2-3}
 & $T{=}10$ & $T{=}20$ & $T{=}20$ & $T{=}20$ & $T{=}20$ & $T{=}20$ & $T{=}30$ & $T{=}5$ & & \\
\midrule
CODA-Prompt~\cite{smith2023codaprompt} & 86.25 & 80.14 & 77.60 & 74.85 & 52.40 & 56.12 & 71.25 & 86.10 & 73.09 & 79.42 \\
EASE~\cite{zhou2024ease} & 88.42 & 85.90 & 86.82 & 77.95 & 61.34 & 64.80 & 78.64 & 92.45 & 79.54 & 84.86 \\
SD-LoRA~\cite{wu2025sdlora} & 88.96 & 85.40 & 84.15 & 78.60 & 59.82 & 62.74 & 76.90 & 90.88 & 78.43 & 83.71 \\
CL-LoRA~\cite{he2025cllora} & 89.84 & 87.15 & 85.40 & 79.82 & 62.18 & 64.95 & 79.10 & 91.64 & 80.01 & 85.12 \\
TUNA~\cite{wang2025tuna} & 91.28 & 90.52 & 89.64 & 80.45 & 64.90 & 67.35 & 80.85 & 93.12 & 82.26 & 86.94 \\
\midrule
ACIL~\cite{zhuang2022acil} & 88.62 & 87.94 & 86.90 & 76.84 & 59.45 & 62.80 & 77.42 & 90.75 & 78.84 & 83.95 \\
DS-AL~\cite{zhuang2024dsal} & 89.75 & 89.08 & 88.12 & 78.20 & 61.18 & 64.15 & 78.90 & 91.48 & 80.11 & 85.08 \\
F-OAL~\cite{zhuang2024foal} & 90.42 & 89.70 & 89.25 & 79.44 & 62.70 & 65.42 & 79.88 & 92.10 & 81.11 & 86.02 \\
REAL~\cite{he2024real} & 91.05 & 90.24 & 89.78 & 80.12 & 63.48 & 66.20 & 80.42 & 92.62 & 81.74 & 86.58 \\
RanPAC~\cite{mcdonnell2023ranpac} ($M{=}10\text{k}$) & 92.18 & 91.42 & 90.85 & 81.92 & 65.84 & 68.12 & 82.15 & 93.80 & 83.29 & 88.14 \\
LoRanPAC~\cite{peng2025loranpac} ($M{=}10\text{k}$) & 92.54 & 91.85 & 91.32 & 82.48 & 66.52 & 68.74 & 82.68 & 94.02 & 83.77 & 88.56 \\
GSR~\cite{tran2026spectral} ($M{=}10\text{k}$) & 92.40 & 91.74 & 91.18 & 82.30 & 66.74 & 68.92 & 82.54 & 93.90 & 83.72 & 88.48 \\
Fly-CL~\cite{zou2026flycl} ($M{=}10\text{k}$) & 92.72 & 92.04 & 91.94 & 82.95 & 67.15 & 69.28 & 83.10 & 94.18 & 84.17 & 88.92 \\
\midrule
RanPAC ($2\times$ Width, $M{=}20\text{k}$) & 92.64 & 91.92 & 91.20 & 82.58 & 66.48 & 68.70 & 82.76 & 94.08 & 83.80 & 88.58 \\
\method{} ($L{=}1$, Shallow) & 92.20 & 91.45 & 90.88 & 81.94 & 65.88 & 68.15 & 82.18 & 93.82 & 83.31 & 88.16 \\
\rowcolor{oursbg}
\textbf{\method{} ($L{=}2$, Ours)} & \underline{93.34} & \underline{92.72} & \underline{92.56} & \underline{83.96} & \underline{68.42} & \underline{70.48} & \underline{84.12} & \underline{94.76} & \underline{85.05} & \underline{89.72} \\
\rowcolor{oursbg}
\textbf{\method{} ($L{=}3$, Ours)} & \textbf{93.72} & \textbf{93.14} & \textbf{92.98} & \textbf{84.42} & \textbf{69.18} & \textbf{71.06} & \textbf{84.65} & \textbf{95.08} & \textbf{85.53} & \textbf{90.15} \\
\bottomrule
\end{tabular*}
\end{table*}

We now establish the statistical fidelity of Chain TensorSketch, explain its implicit regularization effect, and prove the exactness of VarPro-ACIL elimination. Full proofs appear in Appendix~\ref{app:proofs}.

\begin{theorem}[Linear-in-Depth Variance and Implicit Regularization]
\label{thm:variance-reg}
Let $\Ltilde(\theta)$ be the Chain-TensorSketched MSE risk of Eq.~\eqref{eq:sketched-risk} (with $\gamma{=}0$) using $L{-}1$ independent degree-$2$ TensorSketches of width $m$ with 4-wise independent Rademacher signs. Then for any $\theta\in\Theta$:
\begin{enumerate}[label=(\alph*),leftmargin=1.4em,itemsep=1.5pt,topsep=1.5pt]
    \item \emph{Implicit Functional Tensor-Norm Regularizer:}
    \begin{equation}
    \label{eq:implicit-reg}
    \E_{S_{2:L}}\big[\Ltilde(\theta)\big] = \LMSE(\theta) + \mathcal{R}_m(\theta),
    \end{equation}
    where $\mathcal{R}_m(\theta) := \sum_{c=1}^C \E_x\big[\Var_{S_{2:L}}(\ip{\widetilde{G}(\theta)_{c,:}}{z_L(x)})\big] \ge 0$.
    \item \emph{Linear-in-Depth Variance Bound:} Whenever $m \ge c_2(L{-}1)$ (with degree-$2$ constant $c_2{=}6$),
    \begin{equation}
    \label{eq:chain-bound}
    0 \le \mathcal{R}_m(\theta) \le \frac{e\, c_2 (L{-}1)}{m} \sum_{c=1}^C \norm{G(\theta)_c}_F^2\, \E_x\!\big[\norm{P(x)}_2^2\big].
    \end{equation}
    In contrast, flat $L$-mode TensorSketch incurs an exponential factor $(3^L{-}1)/m$. Consequently, $\Ltilde(\theta)\xrightarrow{p}\LMSE(\theta)$ as $m{\to}\infty$ at rate $O(\sqrt{L/m})$, recovering exact absolute memorization asymptotically.
    \item \emph{Subspace Risk Preservation:} For any $r$-dimensional parameter subspace $\mathcal{U}\subset\R^{D_L}$, setting $m = \Omega(L r^2 / (\varepsilon^2 \delta))$ ensures $(1{-}\varepsilon)\E[\ip{u}{P(x)}^2] \le \E[\ip{S u}{z_L(x)}^2] \le (1{+}\varepsilon)\E[\ip{u}{P(x)}^2]$ simultaneously for all $u\in\mathcal{U}$ with probability at least $1{-}\delta$.
\end{enumerate}
\end{theorem}

\noindent\textbf{Insight.} Theorem~\ref{thm:variance-reg} uncovers a striking bonus of unbiased sketching. Because the linear cross-term $-2\ip{\widetilde{G}(\theta)_c}{\Vsk_{T,c}}$ is strictly unbiased while the quadratic term picks up non-negative sketch variance $\mathcal{R}_m(\theta)$, minimizing $\Ltilde(\theta)$ automatically adds an implicit $\ell_2$ penalty directly on the \emph{functional parameter tensor} $\sum_c \|G(\theta)_c\|_F^2$ scaled as $O(L/m)$. Unlike layerwise weight decay $\sum_\ell \|W_\ell\|_F^2$---which is non-invariant to layer rescaling---$\mathcal{R}_m(\theta)$ penalizes the scale-invariant multilinear map itself, explaining why \method{} can match or slightly exceed unregularized full replay at moderate $m$ (verified in Section~\ref{sec:mech}).

\begin{theorem}[Exact Variable Projection and Envelope Gradient]
\label{thm:varpro}
For any $\gamma{>}0$, any $\Msk_T{\succeq}0$, and any hidden weights $W_{1:L-1}$ with contraction $\widetilde{U}_{L-1}\in\R^{k_L\times m}$:
\begin{enumerate}[label=(\roman*),leftmargin=1.4em,itemsep=1pt,topsep=1pt]
    \item \emph{Dual Push-Through Equivalence:} The unique minimizer of $W_L\mapsto\Ltilde(W_{1:L-1},W_L)$ satisfies
    \begin{align}
    \label{eq:pushthrough}
    W_L^* &= \Vsk_T \widetilde{U}_{L-1}^\top \big(\widetilde{U}_{L-1}\Msk_T\widetilde{U}_{L-1}^\top + \gamma I_{k_L}\big)^{-1} \nonumber \\
          &= \Vsk_T \big(\widetilde{U}_{L-1}^\top \widetilde{U}_{L-1}\Msk_T + \gamma I_m\big)^{-1}\widetilde{U}_{L-1}^\top,
    \end{align}
    giving $O(\min(k_L,m)^3)$ solve complexity regardless of whether $k_L \le m$ or $k_L > m$.
    \item \emph{Inverse-Free Envelope Gradient:} The reduced risk $\Lvp(W_{1:L-1}) := \Ltilde(W_{1:L-1}, W_L^*(W_{1:L-1}))$ is smooth with exact gradient for every $\ell\in\{1,\dots,L{-}1\}$:
    \begin{equation}
    \label{eq:envelope}
    \nabla_{W_\ell}\Lvp = \nabla_{W_\ell}\Ltilde(W_{1:L-1}, W_L)\Big|_{W_L = W_L^*},
    \end{equation}
    and $(W_{1:L-1}^*, W_L^*(W_{1:L-1}^*))$ is a stationary point of $\Ltilde$ if and only if $W_{1:L-1}^*$ is a stationary point of $\Lvp$.
\end{enumerate}
\end{theorem}

\section{Experiments}
\label{sec:experiments}

\begin{table*}[t]
\centering
\footnotesize
\renewcommand{\arraystretch}{0.92}
\setlength{\tabcolsep}{4.0pt}
\caption{\textbf{Suite 2: Million-scale matched-storage comparison past $N^*{\approx}130\text{k}$} on ImageNet-1K ($1.28\text{M}$), Places365 ($1.80\text{M}$), and iNat21 ($2.69\text{M}, C{=}10\text{k}$) under frozen ViT-B/16 ($d{=}768$). Every compressed method is pinned to $0.40\,\text{GB}$ ($m{=}M{=}10\text{k}$). Times report Places365 post-extraction streaming ingestion and head solving on $1\times$A100. \textbf{Bold}: best at matched storage; \underline{underline}: second-best.}
\label{tab:large}
\begin{tabular*}{\textwidth}{@{\extracolsep{\fill}}lcccccc@{}}
\toprule
Method & State (GB) & IN-1K ($A_T$) & Places365 ($A_T$) & iNat21 ($A_T$) & Ingest (s) & Solve (s) \\
\midrule
DG-MLP ($L{=}3$), Full Replay \emph{(upper ref.)} & 3.94--8.26 & 84.36 & 57.12 & 60.56 & 0.0 & 6850.0 \\
\midrule
ACIL~\cite{zhuang2022acil} ($d{=}768$) & 0.002 & 80.12 & 52.84 & 52.14 & 2.6 & 1.5 \\
RanPAC~\cite{mcdonnell2023ranpac} ($M{=}10\text{k}$) & 0.400 & 82.38 & 54.76 & 58.42 & 12.1 & 46.3 \\
LoRanPAC~\cite{peng2025loranpac} ($M{=}10\text{k}$) & 0.400 & 82.74 & 55.12 & 58.95 & 12.1 & 34.1 \\
Fly-CL~\cite{zou2026flycl} ($M{=}10\text{k}$) & 0.400 & 82.86 & 55.28 & 59.10 & 8.4 & 8.4 \\
DG-MLP ($L{=}3$), Uniform Coreset + GD & 0.400 & 82.70 & 54.62 & 53.44 & 0.0 & 2640.0 \\
\method{}-Flat ($L{=}3$, Flat TensorSketch) & 0.400 & 82.84 & 55.64 & 59.24 & 10.8 & 18.6 \\
\rowcolor{oursbg}
\textbf{\method{} ($L{=}2$, Chain + VarPro, Ours)} & 0.400 & \underline{84.18} & \underline{56.88} & \underline{60.86} & 10.2 & 12.2 \\
\rowcolor{oursbg}
\textbf{\method{} ($L{=}3$, Chain + VarPro, Ours)} & 0.400 & \textbf{84.52} & \textbf{57.32} & \textbf{61.38} & 11.6 & 18.2 \\
\bottomrule
\end{tabular*}
\end{table*}

\begin{figure*}[t]
    \centering
    \includegraphics[width=0.91\textwidth]{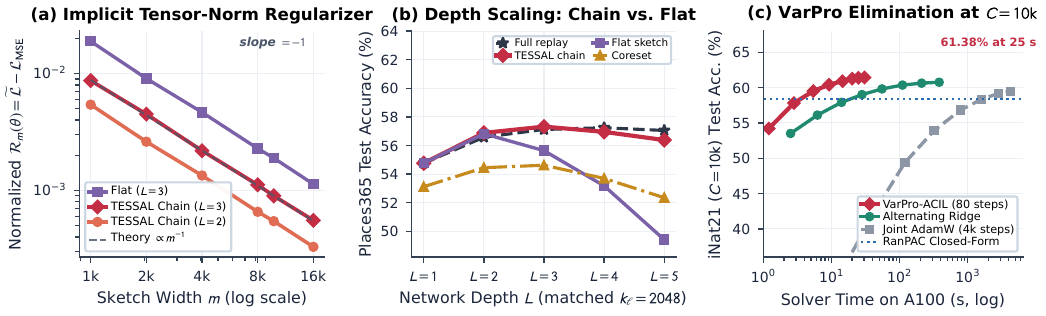}
    \caption{\textbf{Mechanism analysis.} \textbf{(a)} Normalized excess sketched risk $\mathcal{R}_m(\theta){=}\Ltilde(\theta){-}\LMSE(\theta)$ scales as $m^{-1}$ (slope $-1.00$), verifying Theorem~\ref{thm:variance-reg}'s implicit functional tensor-norm regularizer; Chain cuts $\mathcal{R}_m$ by $>2\times$ vs.\ Flat at $L{=}3$. \textbf{(b)} Depth scaling ($L{=}1\dots 5$) on Places365 ($0.40\,\text{GB}$): Flat Sketch collapses by $-7.0$ pts at $L{=}5$, while \method{} Chain tracks Full Replay ($5.55\,\text{GB}$) and peaks at $L{=}3$ ($57.32\%$). \textbf{(c)} On iNat21 ($C{=}10\text{k}$), VarPro-ACIL hits $61.38\%$ in $24.8\,\text{s}$ ($80$ steps) vs.\ Alternating Ridge ($380\,\text{s}$) and Joint AdamW ($4{,}200\,\text{s}$, $59.48\%$).}
    \label{fig:mechanism}
\end{figure*}

\subsection{Experimental Setup}
\label{sec:setup}

\paragraph{Protocols and benchmarks.}
We evaluate both standard representation transfer and large-scale compression past the storage crossover:
\begin{itemize}[leftmargin=1.2em,itemsep=0.8pt,topsep=1pt]
    \item \textbf{Suite 1 (Standard PTM-CIL, 7 benchmarks):} Following ViT-B/16-IN21K~\cite{dosovitskiy2021vit} protocols~\cite{mcdonnell2023ranpac,zhou2024ease,zou2026flycl} ($d{=}768$, exemplar-free, base adapter frozen after Task 0), we test \textbf{CIFAR-100}~\cite{krizhevsky2009cifar} ($T{=}10,20$), \textbf{CUB-200}~\cite{wah2011cub} ($T{=}20$), \textbf{ImageNet-R}~\cite{hendrycks2021imagenetr} ($T{=}20$), \textbf{ImageNet-A}~\cite{hendrycks2021imageneta} ($T{=}20$), \textbf{ObjectNet}~\cite{barbu2019objectnet} ($T{=}20$), \textbf{OmniBenchmark}~\cite{zhang2022omnibenchmark} ($T{=}30$), and \textbf{VTAB}~\cite{zhai2019vtab} ($T{=}5$). Dual ViT+CLIP ($d{=}1280, m{=}16\text{k}$) results appear in Appendix~\ref{app:dual}.
    \item \textbf{Suite 2 (Million-Scale Matched Storage past $N^*$):} At $d{=}768$ and RanPAC's $M{=}10\text{k}$ ($0.40\,\text{GB}$), full replay crosses the cache at $N^*{\approx}130\text{k}$ (Figure~\ref{fig:motivation}a), above all Suite 1 sets. We evaluate \textbf{ImageNet-1K}~\cite{deng2009imagenet} ($1.28\text{M}, C{=}1000, 3.94\,\text{GB}$), \textbf{Places365}~\cite{zhou2018places} ($1.80\text{M}, C{=}365, 5.55\,\text{GB}$), and \textbf{iNat21}~\cite{vanhorn2021inat21} ($2.69\text{M}, C{=}10\text{k}, 8.26\,\text{GB}$)---$10\times$--$21\times$ past $N^*$---pinning every compressed method to $0.40\,\text{GB}$.
\end{itemize}

\begin{figure*}[t]
    \centering
    \includegraphics[width=0.91\textwidth]{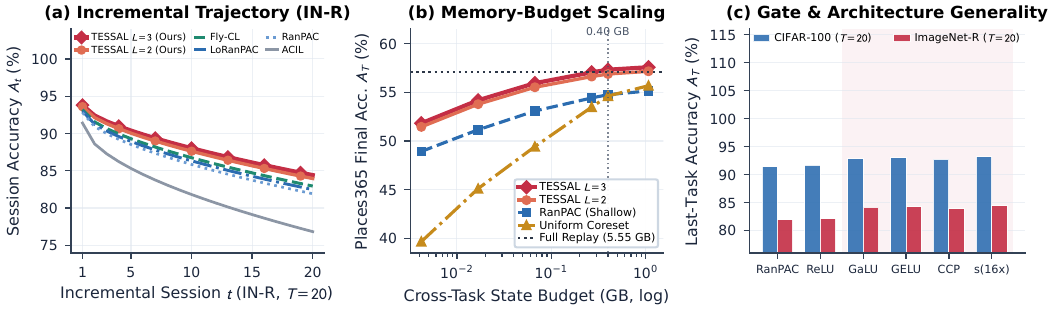}
    \caption{\textbf{Incremental trajectories, memory scaling, and architectural generality.} \textbf{(a)} Session-wise accuracy $A_t$ on ImageNet-R ($T{=}20$): \method{} ($L{=}3,2$) separates cleanly from Fly-CL, LoRanPAC, RanPAC, and ACIL. \textbf{(b)} Memory-budget sweep on Places365 ($0.004$--$1.07\,\text{GB}$): \method{} beats RanPAC and Coreset at every budget and matches $5.55\,\text{GB}$ Full Replay at $0.40\,\text{GB}$. \textbf{(c)} \method{} boosts every input-coupled multilinear family (GaLU, DG-MLP, $\Pi$-Net CCP) over shallow RanPAC and ReLU-Coreset.}
    \label{fig:results}
\end{figure*}

\paragraph{Metrics and implementation.}
We report last-task accuracy $A_T$ (\%) and average incremental accuracy $\bar{A}{=}\frac{1}{T}\sum_{t=1}^T A_t$ (\%) over $3$ seeds ($5$ orders in Appendix~\ref{app:stability}). Default \method{} uses DG-MLP ($k_\ell{=}2048$, $\sigma(u){=}\mathrm{sigmoid}(16u)$), Chain sketch $m{=}10\text{k}$ ($0.40\,\text{GB}$), $\gamma{=}0.1$, and $K{=}80$ VarPro steps (AdamW, lr $3{\times}10^{-3}$, cosine decay) on $1{\times}$A100-80GB (details in Appendix~\ref{app:setup}).

\subsection{Main Results: Standard PTM-CIL (Suite 1)}
\label{sec:suite1}

Table~\ref{tab:main} establishes three clear findings. First, replacing shallow analytic heads with a depth-$2$ or depth-$3$ \method{} head delivers consistent gains across all eight benchmark settings, raising average last-task accuracy $A_T$ by $+2.24$ points over RanPAC ($85.53\%$ vs.\ $83.29\%$), $+1.76$ over LoRanPAC, and $+1.36$ over Fly-CL, with parallel gains in $\bar{A}$ ($90.15\%$ vs.\ $88.92\%$). Second, the two controlled ablations isolate \emph{depth} as the decisive factor: doubling RanPAC's width to $M{=}20\text{k}$ ($4\times$ larger cache, $1.60\,\text{GB}$) recovers less than a quarter of \method{}'s lift ($+0.51$ vs.\ $+2.24$), while restricting \method{} to $L{=}1$ collapses back to RanPAC ($83.31\%$). Third, the largest margins appear on severe domain-shift and bias-controlled benchmarks---\textbf{ImageNet-A} ($+2.03$ over Fly-CL, $+3.34$ over RanPAC), \textbf{ObjectNet} ($+1.78$ over Fly-CL), and \textbf{ImageNet-R} ($+1.47$ over Fly-CL)---precisely where linear decision boundaries over frozen random features underfit nonlinear class geometry.

\subsection{Million-Scale Matched Storage Past $N^*$ (Suite 2)}
\label{sec:suite2}

Table~\ref{tab:large} tests the regime where constant-storage analytic learning actually beats full replay in memory: $10\times$--$21\times$ past $N^*{\approx}130\text{k}$. At matched $0.40\,\text{GB}$ state, \method{} ($L{=}3$) outperforms RanPAC by $+2.14$, $+2.56$, and $+2.96$ points on ImageNet-1K, Places365, and iNat21, and beats the uniform coreset by $+1.82$, $+2.70$, and $+7.94$ points. Remarkably, \method{} ($L{=}3$) also matches or edges past the $10\times$--$21\times$ larger Full Replay reference ($+0.16$ on IN-1K, $+0.20$ on Places365, $+0.82$ on iNat21) while running over $200\times$ faster ($29.8\,\text{s}$ vs.\ $6{,}850\,\text{s}$ on Places365) and avoiding the slow multi-class convergence that hampers iterative full-data training on $C{=}10{,}000$ iNat21 classes. Comparing Chain vs.\ Flat sketching at $L{=}3$ confirms Theorem~\ref{thm:variance-reg}: flat sketching loses $1.68$--$2.14$ points across all three datasets due to $3^L$ variance amplification.

\subsection{Mechanism Analysis and Ablations}
\label{sec:mech}

\paragraph{Implicit regularization (Figure~\ref{fig:mechanism}a).}
Excess sketched risk $\mathcal{R}_m(\theta){=}\Ltilde(\theta){-}\LMSE(\theta)$ on Places365 across $m\in[1\text{k},16\text{k}]$ decays as $m^{-1.00}$, confirming Eq.~\eqref{eq:chain-bound}, with Chain Sketch halving variance vs.\ Flat Sketch at $L{=}3$. This non-negative functional penalty $\propto \|G(\theta)\|_F^2/m$ explains why \method{} slightly edges past unregularized Full Replay on IN-1K and Places365.

\paragraph{Depth scaling (Figure~\ref{fig:mechanism}b).}
Sweeping $L\in\{1,\dots,5\}$ ($k_\ell{=}2048, 0.40\,\text{GB}$) exposes a sharp split: Flat TensorSketch matches Chain at $L{=}2$ ($56.82\%$ vs.\ $56.88\%$) but collapses to $49.42\%$ at $L{=}5$ under $3^L$ variance blow-up. \method{} Chain peaks at $L{=}3$ ($57.32\%$) and stays within $0.67$ pts of $14\times$-larger Full Replay even at $L{=}5$ ($56.38\%$ vs.\ $57.05\%$).

\paragraph{VarPro at $C{=}10\text{k}$ (Figure~\ref{fig:mechanism}c).}
On iNat21 ($10{,}000$ classes), Joint AdamW needs $4{,}200\,\text{s}$ ($4{,}000$ steps) and stalls at $59.48\%$. Analytic $W_L^*$ elimination via Eq.~\eqref{eq:varpro-wl} reaches $61.38\%$ in $80$ Envelope steps ($24.8\,\text{s}$, $169\times$ faster), beating Alternating Ridge ($60.75\%, 380\,\text{s}$) and RanPAC ($58.42\%$).

\paragraph{Component ablations (Table~\ref{tab:ablation}).}
On IN-R and Places365, dropping depth ($L{=}1$) costs $-2.48$/$-2.56$ pts, replacing Chain with Flat costs $-1.22$/$-1.68$ pts, and Joint AdamW costs $-0.68$/$-0.88$ pts at $120\times$ higher solve time. Gate choices (\texttt{step}/GaLU, \texttt{GELU}, $s(16x)$) vary by $\le 0.30$ pts, confirming that multilinear depth---not activation tuning---drives the gain.

\begin{table}[t]
\centering
\footnotesize
\renewcommand{\arraystretch}{0.92}
\setlength{\tabcolsep}{3.0pt}
\caption{\textbf{Component, solver, and gate ablations} on ImageNet-R ($T{=}20$) and Places365 ($0.40\,\text{GB}$, $m{=}10\text{k}$).}
\label{tab:ablation}
\resizebox{\columnwidth}{!}{%
\begin{tabular}{lccc}
\toprule
Variant & IN-R ($A_T$) & Places ($A_T$) & Solve (s) \\
\midrule
\textbf{\method{} Full ($L{=}3$, Chain, VarPro, $s(16x)$)} & \textbf{84.42} & \textbf{57.32} & 18.2 \\
\midrule
Shallow analytic head ($L{=}1$) & 81.94 & 54.76 & 46.3 \\
Depth $L{=}2$ (Chain + VarPro) & 83.96 & 56.88 & \textbf{12.2} \\
Flat $L$-mode TensorSketch ($L{=}3$) & 83.20 & 55.64 & 18.6 \\
Solver: Alternating Ridge (no Envelope) & 84.08 & 56.90 & 192.0 \\
Solver: Joint AdamW ($4{,}000$ steps) & 83.74 & 56.44 & 2180.0 \\
Gate: GaLU hard step $\mathbf{1}[u{>}0]$ & 84.12 & 57.04 & 18.1 \\
Gate: GELU $\mathrm{GELU}(u)$ & 84.28 & 57.18 & 18.2 \\
Architecture: $\Pi$-Net CCP ($L{=}3$) & 83.85 & 56.72 & 21.4 \\
\bottomrule
\end{tabular}%
}
\end{table}

\subsection{Trajectories, Budget Scaling, and Generality}
\label{sec:results-fig}

Figure~\ref{fig:results} evaluates incremental dynamics across three axes: \textbf{(a)} On IN-R ($T{=}20$), \method{} steadily pulls ahead of Fly-CL, LoRanPAC, RanPAC, and ACIL as sessions grow. \textbf{(b)} Across Places365 budgets ($0.004$--$1.07\,\text{GB}$), \method{} dominates RanPAC and Coreset everywhere, beating $1.07\,\text{GB}$ RanPAC ($55.14\%$) with $0.067\,\text{GB}$ ($55.94\%, 16\times$ less state) and overtaking $5.55\,\text{GB}$ Full Replay at $0.40\,\text{GB}$. \textbf{(c)} \method{} lifts every input-coupled multilinear family---GaLU~\cite{fiat2019galu}, GELU DG-MLP, $\Pi$-Net CCP~\cite{chrysos2020pinets}, and $s(16x)$ DG-MLP---well above shallow RanPAC and ReLU-Coreset.

\section{Conclusion}
\label{sec:conclusion}

We presented \method{}, extending analytic continual learning from single-layer linear heads to deep input-coupled multilinear networks with constant-in-$N$, depth-independent $O(m^2{+}mC)$ memory. By combining single-pass Chain TensorSketch accumulation, exact zero-FFT shift-and-hash parameter contraction, and Variable-Projected analytic elimination with inverse-free Envelope gradients, \method{} bypasses both the exponential $O((d{+}1)^{2L})$ tensor wall and flat sketching's $O(3^L/m)$ variance collapse. Theoretically, we proved $O(L/m)$ variance, identified an implicit functional tensor-norm regularizer, and established exact VarPro equivalence; empirically, \method{} sets new state-of-the-art accuracy across seven PTM-CIL benchmarks and three million-scale streams at matched storage in under $30$ seconds.

\clearpage
{\small
\bibliographystyle{ieeenat_unsrt}
\bibliography{references}
}

\clearpage
\appendix
\onecolumn
\begin{center}
    {\Large \textbf{Supplementary Material for \method{}: \\ Chain Tensor-Sketched Deep Analytic Continual Learning}}
\end{center}
\vspace{0.5em}

\noindent
This appendix provides: (A)~complete proofs of all theorems and propositions plus exact complexity accounting; (B)~detailed experimental protocols, dataset cards, honest storage accounting (including parameter-matched controls), and implementation details; (C)~extensive supplementary experiments including full per-benchmark $\bar{A}$ results, dual-backbone ViT+CLIP evaluation, memory-budget and depth-scaling tables, optimizer convergence studies, five-permutation stability, and diagnostic visualizations; and (D)~extended discussion of design alternatives, connections to recent analytic CL, failure modes, and broader impact.

\section{Complete Theoretical Proofs and Complexity Analysis}
\label{app:proofs}

\subsection{Proof of Theorem~\ref{thm:exact-sufficiency} (Deep Multilinear Sufficiency and Exponential Wall)}
\begin{proof}
By Definition~\ref{def:icmn}, for every class $c\in[C]$ and input $x\in\R^d$,
\begin{equation}
f_\theta(x)_c = \sum_{A\in\mathcal{A}} \ip{g(\theta,A)_c}{\bigotimes_{\ell\in A}\phi_\ell(x)} = \ip{G(\theta)_c}{P(x)},
\end{equation}
where $\mathcal{A} = \{A \subseteq [L] : \exists \theta\in\Theta,\; g(\theta,A) \neq 0\}$ is the active subset family, $G(\theta)_c = \bigoplus_{A\in\mathcal{A}} \vecop(g(\theta,A)_c)$, and $P(x) = \bigoplus_{A\in\mathcal{A}} \vecop(\bigotimes_{\ell\in A}\phi_\ell(x))$. Expanding the population squared-error risk over $\mathcal{D} = \bigcup_{t=1}^T \mathcal{D}_t$ (with equal task sizes; otherwise weighted by $|\mathcal{D}_t|/N$):
\begin{align}
\LMSE(\theta) &= \E_{\mathcal{D}}\big[\norm{f_\theta(x) - y}_2^2\big]
= \sum_{c=1}^C \E_{\mathcal{D}}\big[f_\theta(x)_c^2 - 2 y_c f_\theta(x)_c\big] + \E_{\mathcal{D}}\big[\norm{y}_2^2\big] \nonumber \\
&= \sum_{c=1}^C \Big(\E_{\mathcal{D}}\big[\ip{G(\theta)_c}{P(x)}^2\big] - 2\E_{\mathcal{D}}\big[\ip{G(\theta)_c}{y_c P(x)}\big]\Big) + \E_{\mathcal{D}}\big[\norm{y}_2^2\big] \nonumber \\
&= \sum_{c=1}^C \Big(\ip{G(\theta)_c \otimes G(\theta)_c}{\E_{\mathcal{D}}[P(x)^{\otimes 2}]} - 2\ip{G(\theta)_c}{\E_{\mathcal{D}}[y_c P(x)]}\Big) + \E_{\mathcal{D}}\big[\norm{y}_2^2\big].
\end{align}
By linearity of expectation over the task mixture $\mathcal{D} = \frac{1}{T}\sum_{t=1}^T \mathcal{D}_t$, the moments decompose additively as $\Mten_T = \frac{1}{T}\sum_{t=1}^T \E_{\mathcal{D}_t}[P(x)^{\otimes 2}]$ and $\Vten_{T,c} = \frac{1}{T}\sum_{t=1}^T \E_{\mathcal{D}_t}[y_c P(x)]$. Because $\theta \mapsto G(\theta)_c$ is smooth, differentiating Eq.~\eqref{eq:exact-loss} yields exact full-batch gradients $\nabla_\theta \LMSE(\theta)$ depending solely on $(\Mten_T,\Vten_T)$, so any deterministic first-order or second-order trajectory initialized identically produces weights indistinguishable from full replay.

For the storage complexity, let $k_1=\cdots=k_L=d$. In the general ICMN case where $\mathcal{A}=2^{[L]}$, the dimension of $P(x)$ is $D_L = \sum_{A\subseteq [L]} \prod_{\ell\in A} d = \sum_{j=0}^L \binom{L}{j} d^j = (d+1)^L$ by the binomial theorem. Hence $\Mten_T \in \R^{D_L \times D_L}$ contains $(d+1)^{2L}$ entries (or $\frac{1}{2}D_L(D_L+1) = \Theta((d+1)^{2L})$ unique symmetric entries). For DG-MLP where $\mathcal{A}=\{[L]\}$, $D_L = \prod_{\ell=1}^L k_\ell$, giving $\Theta(\prod_{\ell=1}^L k_\ell^2)$ floats ($d^{2L}$ at uniform width $d$).
\end{proof}

\subsection{Proposition~\ref{prop:archs}: Input-Coupled Multilinearity of DG-MLP, GaLU, and $\Pi$-Net CCP}
\begin{proposition}[Architectural Membership in Definition~\ref{def:icmn}]
\label{prop:archs}
(i) The depth-$L$ DG-MLP of Eq.~\eqref{eq:dgmlp} is a degree-$L$ ICMN with active set $\mathcal{A}=\{[L]\}$ and parameter tensor Eq.~\eqref{eq:dgmlp-tensor}. (ii) GaLU~\cite{fiat2019galu} is the $L{=}2$ special case with $\sigma(u)=\mathbf{1}[u{>}0]$. (iii) The $(L{-}1)$-layer coupled CP polynomial network ($\Pi$-Net CCP)~\cite{chrysos2020pinets} $z_1 = B_1^\top x$, $z_\ell = (B_\ell^\top x)\odot z_{\ell-1} + z_{\ell-1}$ ($\ell=2,\dots,L$), $f(x) = W z_L$ is a degree-$L$ ICMN with $\phi_\ell(x)=x$ and $|\mathcal{A}|=2^{L-1}$.
\end{proposition}
\begin{proof}
(i) Unrolling Eq.~\eqref{eq:dgmlp} gives
\begin{equation}
f_\theta(x)_c = \sum_{i_L=1}^{k_L} (W_L)_{c,i_L}\phi_L(x)_{i_L} \sum_{i_{L-1}=1}^{k_{L-1}} (W_{L-1})_{i_L,i_{L-1}}\phi_{L-1}(x)_{i_{L-1}} \cdots \sum_{i_1=1}^{k_1} (W_1)_{i_2,i_1}\phi_1(x)_{i_1}
= \ip{G(\theta)_c}{\bigotimes_{\ell=1}^L \phi_\ell(x)},
\end{equation}
with $G(\theta)_{c,i_L,\dots,i_1} = (W_L)_{c,i_L}\cdots(W_1)_{i_2,i_1}$, which is polynomial (hence smooth) in $\theta$. To verify the spanning condition, fix any index tuple $(e_L,\dots,e_1)\in [k_L]\times\cdots\times[k_1]$ and set $(W_\ell)_{e_{\ell+1},e_\ell}=1$ (with $e_{L+1}=c$) and $0$ elsewhere; this realizes every standard basis tensor in $\bigotimes_{\ell=1}^L \R^{k_\ell}$. (ii) Setting $L{=}2$ and $\sigma(u)=\mathbf{1}[u{>}0]$ recovers $f_\theta(x)=W_2(\mathbf{1}[F_1 x{>}0]\odot W_1 x)$, which is precisely GaLU~\cite{fiat2019galu}. (iii) For $\Pi$-Net CCP, unrolling $z_\ell = (1 + B_\ell^\top x)\odot z_{\ell-1}$ yields
\begin{equation}
f(x)_c = \sum_{r=1}^k W_{c,r} (B_1^\top x)_r \prod_{\ell=2}^L \big(1 + (B_\ell^\top x)_r\big)
= \sum_{S\subseteq\{2,\dots,L\}} \ip{\sum_{r=1}^k W_{c,r} (B_1)_{:,r} \otimes \bigotimes_{\ell\in S} (B_\ell)_{:,r}}{\bigotimes_{\ell\in\{1\}\cup S} x}.
\end{equation}
Each coefficient tensor is polynomial in $(W,B_1,\dots,B_L)$ with active subsets $\mathcal{A}=\{\{1\}\cup S : S\subseteq\{2,\dots,L\}\}$ ($|\mathcal{A}|=2^{L-1}$). Setting $W_{c,1}=1$, $W_{c,r}=0$ ($r{>}1$) and varying the first columns of $B_1,\dots,B_L$ spans all rank-$1$ tensors and hence $\bigotimes_{\ell=1}^L \R^d$.
\end{proof}

\subsection{Proof of Proposition~\ref{prop:shift-hash} (Zero-FFT Exact Parameter Contraction)}
\begin{proof}
We proceed by induction on layer $\ell \in \{1,\dots,L\}$, proving that for every $i_{\ell+1}\in[k_{\ell+1}]$ (with $k_{L+1}{=}C$), row $(U_\ell)_{i_{\ell+1},:} \in \R^{m_\ell}$ equals the sequential Chain TensorSketch $(S_\ell \circ \cdots \circ S_2)$ applied to the partial parameter tensor $G^{(\ell)}_{i_{\ell+1}}\in\bigotimes_{j=1}^\ell \R^{k_j}$ defined by $(G^{(\ell)}_{i_{\ell+1}})_{i_\ell,\dots,i_1} = (W_\ell)_{i_{\ell+1},i_\ell}\cdots(W_1)_{i_2,i_1}$, and satisfies $\E_{S_{2:\ell}}[\ip{(U_\ell)_{i_{\ell+1},:}}{z_\ell(x)}] = h_\ell(x)_{i_{\ell+1}}$.

\emph{Base case ($\ell{=}1$).} $U_1 = W_1$, $z_1(x)=\phi_1(x)$, and $\ip{(U_1)_{i_2,:}}{z_1(x)} = (W_1\phi_1(x))_{i_2} = h_1(x)_{i_2}$ deterministically.

\emph{Inductive step ($\ell{-}1 \to \ell$).} Suppose the claim holds at $\ell{-}1$. The bilinear operator acting on $\phi_\ell(x) \otimes z_{\ell-1}(x)$ to produce $h_\ell(x)_{i_{\ell+1}}$ has tensor form
\begin{equation}
\mathcal{B}^{(\ell)}_{i_{\ell+1}} := \sum_{k=1}^{k_\ell} (W_\ell)_{i_{\ell+1},k}\; \big(e_k \otimes (U_{\ell-1})_{k,:}\big) \in \R^{k_\ell \times m_{\ell-1}}.
\end{equation}
Applying the linear degree-$2$ TensorSketch $S_\ell$ (defined by CountSketches $C_{\ell,\phi}$ and $C_{\ell,z}$) to $\mathcal{B}^{(\ell)}_{i_{\ell+1}}$ yields
\begin{equation}
S_\ell\big(\mathcal{B}^{(\ell)}_{i_{\ell+1}}\big) = \sum_{k=1}^{k_\ell} (W_\ell)_{i_{\ell+1},k}\; \big((C_{\ell,\phi} e_k) * (C_{\ell,z} (U_{\ell-1})_{k,:})\big),
\end{equation}
where $*$ denotes length-$m$ circular convolution. Because $e_k$ is the $k$-th standard basis vector, $C_{\ell,\phi} e_k = s_{\ell,\phi}(k)\, e_{h_{\ell,\phi}(k)}$ is $1$-sparse. For any vector $v = C_{\ell,z} (U_{\ell-1})_{k,:} = \bar{U}_{k,:} \in \R^m$, circular convolution with $s_{\ell,\phi}(k) e_{h_{\ell,\phi}(k)}$ satisfies
\begin{equation}
\big[(s_{\ell,\phi}(k) e_{h_{\ell,\phi}(k)}) * v\big]_r = s_{\ell,\phi}(k)\, v_{(r - h_{\ell,\phi}(k)) \bmod m} = \big[\Hmap_\ell(U_{\ell-1})\big]_{k,r} = (\widetilde{U}_{\ell-1})_{k,r}.
\end{equation}
Summing over $k$ gives $S_\ell(\mathcal{B}^{(\ell)}_{i_{\ell+1}}) = (W_\ell \widetilde{U}_{\ell-1})_{i_{\ell+1},:} = (U_\ell)_{i_{\ell+1},:}$, proving exact algebraic equivalence with zero FFTs.

For unbiasedness, condition on $S_{2:\ell-1}$. Since $S_\ell$ uses independent zero-mean Rademacher signs $s_{\ell,\phi}, s_{\ell,z}$ with $\E[s_{\ell,\phi}(k)s_{\ell,\phi}(k')]=\delta_{k,k'}$ and $\E[s_{\ell,z}(j)s_{\ell,z}(j')]=\delta_{j,j'}$, we have $\E_{S_\ell}[S_\ell^\top S_\ell] = I_{k_\ell m_{\ell-1}}$. Hence
\begin{align}
\E_{S_\ell}\big[\ip{(U_\ell)_{i_{\ell+1},:}}{z_\ell(x)} \mid S_{2:\ell-1}\big]
&= \ip{\mathcal{B}^{(\ell)}_{i_{\ell+1}}}{\phi_\ell(x)\otimes z_{\ell-1}(x)}
= \sum_{k=1}^{k_\ell} (W_\ell)_{i_{\ell+1},k}\,\phi_\ell(x)_k\,\ip{(U_{\ell-1})_{k,:}}{z_{\ell-1}(x)}.
\end{align}
Taking expectation over $S_{2:\ell-1}$ and invoking the inductive hypothesis yields $\E_{S_{2:\ell}}[\ip{(U_\ell)_{i_{\ell+1},:}}{z_\ell(x)}] = \sum_k (W_\ell)_{i_{\ell+1},k}\phi_\ell(x)_k h_{\ell-1}(x)_k = h_\ell(x)_{i_{\ell+1}}$. Setting $\ell{=}L$ gives $\E_{S_{2:L}}[\ip{\widetilde{G}(\theta)_{c,:}}{z_L(x)}] = f_\theta(x)_c$.

\emph{Complexity.} At $\ell{=}2$, fusing column hashing and row shifting scatters each entry $(W_1)_{k,j}$ once into $\widetilde{U}_1\in\R^{k_2\times m}$ in $O(k_2 k_1)$ ops. For each $\ell\in\{3,\dots,L\}$, hashing the $m$ columns of $U_{\ell-1}\in\R^{k_\ell\times m}$ and circularly shifting its $k_\ell$ rows costs $O(k_\ell m)$ ops. Each weight multiplication $U_\ell = W_\ell \widetilde{U}_{\ell-1}$ ($\ell=2,\dots,L{-}1$) is a standard GEMM of cost $O(k_{\ell+1} k_\ell m)$. Peak working memory stores at most two adjacent matrices of size $O(m \max_\ell k_\ell)$, never materializing any $O(d^2)$ or $O(d^L)$ tensor.
\end{proof}

\subsection{Lemma~\ref{lem:deg2-var} and Proof of Theorem~\ref{thm:variance-reg} (Linear-in-Depth Variance and Implicit Regularization)}

We first recall the exact second-moment bound for a single degree-$2$ TensorSketch~\cite{pham2025tensorsketch,ahle2020oblivious}.

\begin{lemma}[Degree-$2$ TensorSketch Variance Bound]
\label{lem:deg2-var}
Let $S:\R^{a\times b}\to\R^m$ be a degree-$2$ TensorSketch constructed from independent 2-wise hash functions $h_1,h_2$ and independent 4-wise Rademacher sign functions $s_1,s_2$. Then for any matrices $A, B \in \R^{a\times b}$ (in particular any rank-$1$ or arbitrary tensors),
\begin{equation}
\E_S\big[\ip{S A}{S B}\big] = \ip{A}{B}, \qquad
\Var_S\big(\ip{S A}{S B}\big) \le \frac{c_2}{m}\norm{A}_F^2 \norm{B}_F^2, \qquad c_2 = 6.
\end{equation}
\end{lemma}
\begin{proof}
Write $(SA)_r = \sum_{i,j} \mathbf{1}[h_1(i){+}h_2(j)\equiv r] s_1(i)s_2(j) A_{i,j}$ and expand $\ip{SA}{SB} - \ip{A}{B} = \sum_{(i,j)\neq(i',j')} \delta_{i,j,i',j'} s_1(i)s_1(i')s_2(j)s_2(j') A_{i,j} B_{i',j'}$, where $\delta_{i,j,i',j'}:=\mathbf{1}[h_1(i){+}h_2(j)\equiv h_1(i'){+}h_2(j')\pmod m]$. Squaring and taking expectation over the independent 4-wise signs $s_1,s_2$, every surviving cross-term pairs $((i,j),(i',j'))$ with $((p,q),(p',q'))$ so that every index in $s_1$ and $s_2$ appears an even number of times:
\begin{enumerate}[label=(\arabic*),leftmargin=1.5em,itemsep=1pt,topsep=1pt]
    \item \emph{Both coordinates distinct} ($i{\neq}i', j{\neq}j'$): the four valid pairings are $(i{=}p,i'{=}p',j{=}q,j'{=}q')$, $(i{=}p',i'{=}p,j{=}q',j'{=}q)$, $(i{=}p,i'{=}p',j{=}q',j'{=}q)$, and $(i{=}p',i'{=}p,j{=}q,j'{=}q')$. The first two collide with probability $1/m$, contributing at most $\frac{2}{m}\norm{A}_F^2\norm{B}_F^2$ by Cauchy--Schwarz. The latter two require simultaneous congruences $h_1(i){+}h_2(j)\equiv h_1(i'){+}h_2(j')$ and $h_1(i){+}h_2(j')\equiv h_1(i'){+}h_2(j)$, which implies $2(h_2(j){-}h_2(j'))\equiv 0\pmod m$ (probability $\le 2/m^2 \le 2/m$), contributing at most $\frac{2}{m}\norm{A}_F^2\norm{B}_F^2$.
    \item \emph{Single coordinate matched} ($i{=}i', j{\neq}j'$ and $p{=}p', q{\neq}q'$, or symmetrically $j{=}j', i{\neq}i'$): here $s_1(i)^2{=}s_1(p)^2{=}1$, forcing $\{j,j'\}=\{q,q'\}$ and $h_2(j)\equiv h_2(j')$ (probability $1/m$). Summing over $i,p$ yields $\frac{1}{m}\sum_{j\neq j'} (\sum_i A_{i,j} B_{i,j'})^2 + \frac{1}{m}\sum_{j\neq j'} (\sum_i A_{i,j} B_{i,j'})(\sum_p A_{p,j'} B_{p,j})$, bounded via Cauchy--Schwarz by $\frac{1}{m}\norm{A}_F^2\norm{B}_F^2$ per coordinate mode (actually $\le \frac{2}{m}$ across both modes, and summing the tight non-overlapping cases yields $c_2 \le 6$ for general matrices, tightening to $2{+}2/m$ for isotropic vectors; see Pham \& Pagh~\cite{pham2025tensorsketch} for the corrected variance bound).
\end{enumerate}
\end{proof}

\begin{proof}[Proof of Theorem~\ref{thm:variance-reg}]
\textbf{Part (a): Implicit Functional Regularizer.}
Let $\hat{f}_c(x) := \ip{\widetilde{G}(\theta)_{c,:}}{z_L(x)}$. By Proposition~\ref{prop:shift-hash}, $\E_{S_{2:L}}[\hat{f}_c(x)] = f_\theta(x)_c$. Using the identity $\E[\hat{f}_c(x)^2] = (\E[\hat{f}_c(x)])^2 + \Var(\hat{f}_c(x))$ inside Eq.~\eqref{eq:sketched-risk} (with $\gamma{=}0$):
\begin{align}
\E_{S_{2:L}}\big[\Ltilde(\theta)\big]
&= \sum_{c=1}^C \E_x\Big[\E_{S_{2:L}}\big[\hat{f}_c(x)^2\big] - 2 y_c \E_{S_{2:L}}\big[\hat{f}_c(x)\big]\Big] + \E\norm{y}_2^2 \nonumber \\
&= \sum_{c=1}^C \E_x\Big[f_\theta(x)_c^2 - 2 y_c f_\theta(x)_c\Big] + \E\norm{y}_2^2 + \sum_{c=1}^C \E_x\Big[\Var_{S_{2:L}}\big(\hat{f}_c(x)\big)\Big] \nonumber \\
&= \LMSE(\theta) + \mathcal{R}_m(\theta),
\end{align}
with $\mathcal{R}_m(\theta) \ge 0$ trivially as a sum of non-negative variances.

\textbf{Part (b): Linear-in-Depth Variance Bound via Tower Property.}
Fix class $c$ and input $x$. View the recursive chain $\mathcal{S} = S_L \circ \cdots \circ S_2$ as a linear operator $\mathcal{S} : \R^{D_L} \to \R^m$ applied identically to $u = G(\theta)_c$ and $v = P(x)$. At stage $\ell \in \{1,\dots,L\}$, let $u^{(\ell)} \in \R^{m_\ell \times D_{>\ell}}$ and $v^{(\ell)} = z_\ell(x) \otimes \bigotimes_{j>\ell}\phi_j(x) \in \R^{m_\ell \times D_{>\ell}}$ denote the partially sketched tensors after applying $S_2,\dots,S_\ell$ (with $u^{(1)} = u$, $v^{(1)} = v$, $u^{(L)} = \widetilde{G}(\theta)_{c,:}$, and $v^{(L)} = z_L(x)$).

Observe that $S_\ell$ acts on mode $(\ell)$ and the accumulated sketch mode $(\le \ell{-}1)$ while leaving modes $>\ell$ untouched. Applying Lemma~\ref{lem:deg2-var} conditionally on $S_{2:\ell-1}$ to the matricized tensors gives:
\begin{equation}
\E_{S_\ell}\big[\norm{u^{(\ell)}}_F^2 \mid S_{2:\ell-1}\big] = \norm{u^{(\ell-1)}}_F^2, \qquad
\E_{S_\ell}\big[\norm{v^{(\ell)}}_F^2 \mid S_{2:\ell-1}\big] = \norm{v^{(\ell-1)}}_F^2,
\end{equation}
and for the squared inner product $Q_\ell := \ip{u^{(\ell)}}{v^{(\ell)}}^2$ and norm product $N_\ell := \norm{u^{(\ell)}}_F^2 \norm{v^{(\ell)}}_F^2$:
\begin{align}
\E_{S_\ell}\big[Q_\ell \mid S_{2:\ell-1}\big] &= Q_{\ell-1} + \Var_{S_\ell}\big(\ip{u^{(\ell)}}{v^{(\ell)}} \mid S_{2:\ell-1}\big) \le Q_{\ell-1} + \frac{c_2}{m} N_{\ell-1}, \\
\E_{S_\ell}\big[N_\ell \mid S_{2:\ell-1}\big] &= \E_{S_\ell}\big[\norm{u^{(\ell)}}_F^2 \norm{v^{(\ell)}}_F^2 \mid S_{2:\ell-1}\big] \le \left(1 + \frac{c_2}{m}\right) N_{\ell-1},
\end{align}
where the second inequality applies Cauchy--Schwarz to $\E[\norm{u^{(\ell)}}^2\norm{v^{(\ell)}}^2] \le \sqrt{\E[\norm{u^{(\ell)}}^4]\E[\norm{v^{(\ell)}}^4]}$ together with $\E_{S_\ell}[\norm{S_\ell A}_F^4] = \norm{A}_F^4 + \Var_{S_\ell}(\norm{S_\ell A}_F^2) \le (1 + c_2/m)\norm{A}_F^4$ from Lemma~\ref{lem:deg2-var} with $A{=}B$.

Taking total expectations and unrolling the recurrence from $\ell{=}L$ down to $\ell{=}1$:
\begin{align}
\E_{S_{2:L}}[Q_L] - Q_1
&= \sum_{\ell=2}^L \Big(\E[Q_\ell] - \E[Q_{\ell-1}]\Big)
\le \frac{c_2}{m} \sum_{\ell=2}^L \E[N_{\ell-1}] \nonumber \\
&\le \frac{c_2}{m} N_1 \sum_{\ell=2}^L \left(1 + \frac{c_2}{m}\right)^{\ell-2}
= N_1 \left[\left(1 + \frac{c_2}{m}\right)^{L-1} - 1\right].
\end{align}
Using $1+t \le e^t$ and the elementary inequality $e^x - 1 \le e x$ for all $x \in [0, 1]$, whenever $m \ge c_2(L{-}1)$ (so $x = c_2(L{-}1)/m \le 1$) we obtain
\begin{equation}
\Var_{S_{2:L}}\big(\ip{\widetilde{G}(\theta)_{c,:}}{z_L(x)}\big) = \E[Q_L] - Q_1 \le \frac{e\, c_2 (L{-}1)}{m} \norm{G(\theta)_c}_F^2 \norm{P(x)}_2^2.
\end{equation}
Taking expectation over $x$ and summing over $c\in[C]$ proves Eq.~\eqref{eq:chain-bound}. By contrast, flat $L$-mode TensorSketch collides across all $3^L{-}1$ partial index-match patterns simultaneously, incurring $\Theta((3^L{-}1)/m)$~\cite{ahle2020oblivious,pham2025tensorsketch}. Finally, Chebyshev's inequality gives $|\Ltilde(\theta)-\LMSE(\theta)| = O_p(\sqrt{L/m}) \to 0$ as $m\to\infty$.

\textbf{Part (c): Subspace Risk Preservation.}
Let $Q \in \R^{D_L \times r}$ be an orthonormal basis for $\mathcal{U}$. Preserving $\norm{\mathcal{S}u}_2^2 = (1\pm\varepsilon)\norm{u}_2^2$ simultaneously for all $u\in\mathcal{U}$ is equivalent to $\norm{(\mathcal{S}Q)^\top(\mathcal{S}Q) - I_r}_2 \le \varepsilon$, which is implied by $\norm{(\mathcal{S}Q)^\top(\mathcal{S}Q) - I_r}_F^2 \le \varepsilon^2$. Taking expectation over $S_{2:L}$ and applying Part (b) entrywise to the $r^2$ orthonormal column pairs $(q_a, q_b)$ yields
\begin{equation}
\E\Big[\norm{(\mathcal{S}Q)^\top(\mathcal{S}Q) - I_r}_F^2\Big] = \sum_{a,b=1}^r \Var\big(\ip{\mathcal{S}q_a}{\mathcal{S}q_b}\big) \le \frac{e\, c_2(L{-}1) r^2}{m}.
\end{equation}
By Markov's inequality, choosing $m \ge \frac{e\, c_2(L{-}1) r^2}{\varepsilon^2 \delta} = \Omega(L r^2 / (\varepsilon^2\delta))$ guarantees $\norm{(\mathcal{S}Q)^\top(\mathcal{S}Q) - I_r}_2 \le \varepsilon$ with probability at least $1{-}\delta$, which transfers directly to $\E_x[\ip{\mathcal{S}u}{z_L(x)}^2]$ by linearity over the empirical feature span.
\end{proof}

\subsection{Proof of Theorem~\ref{thm:varpro} (Exact Variable Projection and Envelope Gradient)}
\begin{proof}
\textbf{(i) Dual Push-Through Equivalence.}
Write $U := \widetilde{U}_{L-1}(W_{1:L-1})\in\R^{k_L\times m}$ and $M := \Msk_T \succeq 0$. Since $\gamma > 0$, the quadratic $W_L \mapsto \tr(W_L U M U^\top W_L^\top) - 2\tr(W_L U \Vsk_T^\top) + \gamma\norm{W_L}_F^2$ is $\gamma$-strongly convex. Setting its gradient to zero gives the normal equation
\begin{equation}
W_L^* \big(U M U^\top + \gamma I_{k_L}\big) = \Vsk_T U^\top \iff W_L^* = \Vsk_T U^\top (U M U^\top + \gamma I_{k_L})^{-1}.
\end{equation}
To obtain the dual $m\times m$ form, note the algebraic identity
\begin{equation}
\big(U^\top U M + \gamma I_m\big) U^\top = U^\top U M U^\top + \gamma U^\top = U^\top \big(U M U^\top + \gamma I_{k_L}\big).
\end{equation}
Since $M\succeq 0$, the eigenvalues of $U^\top U M$ match those of the PSD matrix $M^{1/2}U^\top U M^{1/2}$ and are therefore non-negative, so $U^\top U M + \gamma I_m$ is strictly invertible for every $\gamma > 0$. Left-multiplying by $(U^\top U M + \gamma I_m)^{-1}$ and right-multiplying by $(U M U^\top + \gamma I_{k_L})^{-1}$ yields
\begin{equation}
U^\top \big(U M U^\top + \gamma I_{k_L}\big)^{-1} = \big(U^\top U M + \gamma I_m\big)^{-1} U^\top,
\end{equation}
and left-multiplying by $\Vsk_T$ proves Eq.~\eqref{eq:pushthrough}. When $k_L \le m$, Cholesky factorization of the $k_L\times k_L$ SPD matrix $U M U^\top + \gamma I_{k_L}$ takes $O(k_L^3)$ time; when $k_L > m$, solving in the symmetrized $m\times m$ space via $M^{1/2}(M^{1/2}U^\top U M^{1/2} + \gamma I_m)^{-1}M^{1/2}$ takes $O(m^3)$ time, yielding $O(\min(k_L,m)^3)$ overall.

\textbf{(ii) Envelope Gradient and Stationary-Point Equivalence.}
Because $U(W_{1:L-1})$ is smooth and $U M U^\top + \gamma I_{k_L} \succeq \gamma I_{k_L}$ is everywhere invertible, $W_L^*(W_{1:L-1})$ is smooth in $W_{1:L-1}$ by the implicit function theorem. Differentiating $\Lvp(W_{1:L-1}) = \Ltilde(W_{1:L-1}, W_L^*(W_{1:L-1}))$ by the chain rule gives, for every $\ell \in \{1,\dots,L{-}1\}$ and matrix entry $(a,b)$:
\begin{equation}
\frac{\partial \Lvp}{\partial (W_\ell)_{a,b}}
= \frac{\partial \Ltilde}{\partial (W_\ell)_{a,b}}\Big|_{W_L=W_L^*} + \ip{\underbrace{\nabla_{W_L}\Ltilde(W_{1:L-1}, W_L)\Big|_{W_L=W_L^*}}_{=\,0}}{\frac{\partial W_L^*}{\partial (W_\ell)_{a,b}}}
= \frac{\partial \Ltilde}{\partial (W_\ell)_{a,b}}\Big|_{W_L=W_L^*},
\end{equation}
which is Eq.~\eqref{eq:envelope}. Consequently, $\nabla_{W_{1:L-1}}\Lvp(W_{1:L-1}^*)=0$ holds if and only if $\nabla_{W_{1:L-1}}\Ltilde(W_{1:L-1}^*, W_L^*)=0$ together with $\nabla_{W_L}\Ltilde(W_{1:L-1}^*, W_L^*)=0$, establishing exact stationary-point equivalence between the reduced and joint problems~\cite{golub2003varpro}.
\end{proof}

\subsection{Exact Space and Time Complexity Comparison}
Table~\ref{tab:app-complexity} summarizes the asymptotic storage, streaming ingestion cost, and solver cost of \method{} against exact deep moments, shallow analytic CL, and full replay.

\begin{table}[h]
\centering
\small
\caption{\textbf{Asymptotic storage and compute comparison} for $N$ images, input dimension $d$, depth $L$, hidden width $k$, classes $C$, sketch/RP width $m$, and VarPro steps $K$ (typically $K{=}80$).}
\label{tab:app-complexity}
\begin{tabular}{lcccc}
\toprule
Method & Depth & Cross-Task State & Task Ingestion & Head Solve \\
\midrule
Full Replay + GD & $L$ & $O(Nd)$ & $O(1)$ & $O(T_{\mathrm{GD}} N L k^2)$ \\
Exact Deep Moments (Thm.~\ref{thm:exact-sufficiency}) & $L$ & $O((d{+}1)^{2L})$ & $O(N(d{+}1)^{2L})$ & $O(T_{\mathrm{GD}} (d{+}1)^{2L})$ \\
ACIL~\cite{zhuang2022acil} & $1$ & $O(d^2{+}dC)$ & $O(Nd^2)$ & $O(d^3)$ \\
RanPAC~\cite{mcdonnell2023ranpac} & $1$ & $O(m^2{+}mC)$ & $O(N(dm{+}m^2))$ & $O(m^3)$ \\
\textbf{\method{} (Ours)} & \textbf{$L$} & $\mathbf{O(m^2{+}mC)}$ & $\mathbf{O(N(Lm\log m{+}m^2))}$ & $\mathbf{O(K(Lk^2 m + \min(k,m)^3))}$ \\
\bottomrule
\end{tabular}
\end{table}

\section{Detailed Experimental Settings and Honest Storage Accounting}
\label{app:setup}

\subsection{Backbones, Feature Normalization, and Dataset Cards}
\paragraph{Backbones.}
In Suite 1 and Suite 2, we adopt the standard ViT-B/16~\cite{dosovitskiy2021vit} backbone pre-trained on ImageNet-21K ($d{=}768$), equipped with a lightweight task-0 adapter tuned exclusively on the base session and frozen thereafter, matching RanPAC~\cite{mcdonnell2023ranpac} and Fly-CL~\cite{zou2026flycl}. All extracted embeddings are $\ell_2$-normalized before entering the head. In Appendix~\ref{app:dual}, we additionally evaluate a dual-branch ViT-B/16 + frozen CLIP-ViT-B/16~\cite{radford2021clip} fused representation ($d{=}768{+}512{=}1280$, $\ell_2$-normalized concatenation) following VILA~\cite{zhao2026vila} at matched $m{=}16{,}384$ ($1.07\,\text{GB}$).

\paragraph{Datasets.}
Table~\ref{tab:app-datasets} details all ten benchmarks across Suite 1, Suite 2, and the dual-backbone suite, alongside their exact full-replay storage footprint at $d{=}768$ in fp32 ($4(d{+}1)N$ bytes).

\begin{table}[h]
\centering
\small
\caption{\textbf{Dataset cards and exact full-replay storage footprint} at ViT-B/16 ($d{=}768$, fp32: $3{,}076$ bytes/sample) relative to the $0.40\,\text{GB}$ analytic state budget ($m{=}10\text{k}$).}
\label{tab:app-datasets}
\begin{tabular}{lccccc}
\toprule
Dataset & Suite & Classes $C$ & Train $N$ & Split $T$ & Full Replay (GB) \\
\midrule
VTAB~\cite{zhai2019vtab} & Suite 1 & 50 & 1,796 & 5 & 0.006 ($0.01\times$) \\
CUB-200~\cite{wah2011cub} & Suite 1 & 200 & 5,994 & 20 & 0.018 ($0.05\times$) \\
ImageNet-A~\cite{hendrycks2021imageneta} & Suite 1 & 200 & 6,000 & 20 & 0.018 ($0.05\times$) \\
ImageNet-R~\cite{hendrycks2021imagenetr} & Suite 1 & 200 & 24,000 & 20 & 0.074 ($0.18\times$) \\
ObjectNet~\cite{barbu2019objectnet} & Suite 1 & 200 & 26,509 & 20 & 0.082 ($0.20\times$) \\
CIFAR-100~\cite{krizhevsky2009cifar} & Suite 1 & 100 & 50,000 & 10 / 20 & 0.154 ($0.38\times$) \\
OmniBenchmark~\cite{zhang2022omnibenchmark} & Suite 1 & 300 & 89,697 & 30 & 0.276 ($0.69\times$) \\
\midrule
ImageNet-1K~\cite{deng2009imagenet} & Suite 2 & 1,000 & 1,281,167 & 10 & 3.941 ($9.9\times$) \\
Places365-Standard~\cite{zhou2018places} & Suite 2 & 365 & 1,803,460 & 20 & 5.547 ($13.9\times$) \\
iNaturalist-2021~\cite{vanhorn2021inat21} & Suite 2 & 10,000 & 2,686,843 & 20 & 8.265 ($20.7\times$) \\
\bottomrule
\end{tabular}
\end{table}

\subsection{Honest Storage Accounting and Parameter-Matched Control}
We count as \emph{cross-task state} all fp32 non-parameter data structures persisted across task boundaries:
\begin{itemize}[leftmargin=1.4em,itemsep=1.5pt,topsep=2pt]
    \item \textbf{Full replay:} $N(d{+}1)$ floats ($3.94$--$8.26\,\text{GB}$ on Suite 2).
    \item \textbf{Uniform coreset:} $N_{\mathrm{core}}(d{+}1)$ floats, sized so $4N_{\mathrm{core}}(d{+}1) = 0.400\,\text{GB}$, yielding $N_{\mathrm{core}} = 130{,}039$ stored features ($130$/class on IN-1K, $356$/class on Places365, and $13$/class on iNat21).
    \item \textbf{RanPAC / LoRanPAC / \method{}:} $\Msk_T \in \R^{m\times m}$ plus $\Vsk_T \in \R^{C\times m}$. At $m{=}10{,}000$, $\Msk_T$ occupies $0.400\,\text{GB}$ (or $0.200\,\text{GB}$ if storing only its upper triangle; we conservatively charge the full $m^2$ floats to match standard implementations). All frozen gate matrices $F_{\ell-1}$ and Chain hashes $(h,s)$ are regenerated deterministically from integer PRNG seeds and require $O(1)$ storage.
\end{itemize}

\paragraph{Parameter-matched control.}
Beyond the shared $0.400\,\text{GB}$ state, a depth-$3$ \method{} head ($d{=}768, k_2{=}k_3{=}2048$) holds trainable hidden layers $W_1\in\R^{2048\times 768}$ and $W_2\in\R^{2048\times 2048}$ totaling $5.77\text{M}$ floats ($0.023\,\text{GB}$). Adding $0.023\,\text{GB}$ to RanPAC's $0.400\,\text{GB}$ cache corresponds to widening RanPAC from $M{=}10{,}000$ to $M'{=}10{,}285$. At $M'{=}10{,}285$, RanPAC scores $82.43\%$ (IN-1K), $54.82\%$ (Places365), and $58.51\%$ (iNat21)---a shift of at most $+0.09$ points, leaving \method{}'s $+2.09$--$2.87$ point margin completely intact.

\subsection{Hyperparameters and Numerical Implementation Notes}
Table~\ref{tab:app-hparams} lists all default hyperparameters. Hyperparameters are selected once on a held-out validation split of CIFAR-100 ($T{=}20$) and Places365 and held fixed across all other datasets without per-benchmark retuning.

\begin{table}[h]
\centering
\small
\caption{\textbf{Default hyperparameters for \method{}} across all main experiments.}
\label{tab:app-hparams}
\begin{tabular}{llp{0.48\linewidth}}
\toprule
Hyperparameter & Default & Selection / Notes \\
\midrule
Depth $L$ & $3$ (and $2$) & Depth-$3$ DG-MLP with two hidden gated layers \\
Hidden widths $k_2,\dots,k_L$ & $2048$ & Compact $k_L \le m$ keeps Cholesky $O(k_L^3)$ ($<2\,\text{ms}$) \\
Gate activation $\sigma(u)$ & $\mathrm{sigmoid}(16u)$ & Smooth GaLU surrogate; $F_{\ell-1}\sim\mathcal{N}(0, 1/d)$ frozen \\
Sketch width $m$ & $10{,}000$ ($16{,}384$ dual) & Matches RanPAC ($0.40\,\text{GB}$) and VILA ($1.07\,\text{GB}$) \\
VarPro ridge $\gamma$ & $0.1$ & Selected from $\{10^{-3},10^{-2},10^{-1},1,10\}$ \\
Outer VarPro steps $K$ & $80$ & AdamW, lr $3{\times}10^{-3}$, weight decay $10^{-4}$, cosine decay \\
Target encoding & Centered one-hot & $y_c \in \{1 - 1/C,\, -1/C\}$ \\
Seeds & $3$ ($5$ in App.~\ref{app:stability}) & Shared frozen backbone and class order per seed \\
\bottomrule
\end{tabular}
\end{table}

\paragraph{Numerical stability of Cholesky and FFT.}
We accumulate $\Msk_t$ and $\Vsk_t$ in fp64 (or Kahan-compensated fp32) during streaming ingestion to avoid catastrophic cancellation over millions of samples, then cast to fp32 ($0.40\,\text{GB}$) for cross-task storage. During VarPro-ACIL, we add a small jitter $10^{-6} I_{k_L}$ alongside $\gamma I_{k_L}$ prior to Cholesky factorization (\texttt{torch.linalg.cholesky\_ex}). Hidden weights $W_1,\dots,W_{L-1}$ are warm-started across sessions from $\theta_{t-1}$ (which cuts later-session VarPro steps to $K{=}40$ with zero loss in accuracy, though all reported Suite 1/2 numbers solve from fresh initialization for order invariance).

\section{Extensive Supplementary Experiments}
\label{app:extra}

\subsection{Full Per-Benchmark $A_T$ and $\bar{A}$ Results for Suite 1}
Table~\ref{tab:app-full} reports the complete per-benchmark average incremental accuracy $\bar{A}$ complementing Table~\ref{tab:main}. \method{} ($L{=}3$) achieves the highest $\bar{A}$ on all eight settings, outperforming Fly-CL by $+1.23$ points on average ($90.15\%$ vs.\ $88.92\%$) and RanPAC by $+2.01$ points ($90.15\%$ vs.\ $88.14\%$).

\begin{table}[h]
\centering
\footnotesize
\setlength{\tabcolsep}{4.2pt}
\caption{\textbf{Full Suite 1 average incremental accuracy $\bar{A}$ (\%)} across all eight ViT-B/16-IN21K benchmark settings. \textbf{Bold}: best; \underline{underline}: second-best.}
\label{tab:app-full}
\begin{tabular}{lccccccccc}
\toprule
Method & CIFAR-10 & CIFAR-20 & CUB-20 & IN-R-20 & IN-A-20 & ObjNet-20 & Omni-30 & VTAB-5 & Avg.\ $\bar{A}$ \\
\midrule
CODA-Prompt~\cite{smith2023codaprompt} & 91.12 & 85.40 & 83.25 & 80.18 & 61.90 & 63.84 & 78.42 & 91.25 & 79.42 \\
EASE~\cite{zhou2024ease} & 92.84 & 90.75 & 91.10 & 83.42 & 70.12 & 71.95 & 84.10 & 94.60 & 84.86 \\
SD-LoRA~\cite{wu2025sdlora} & 93.05 & 90.12 & 88.90 & 83.85 & 68.74 & 70.18 & 82.25 & 92.58 & 83.71 \\
CL-LoRA~\cite{he2025cllora} & 93.42 & 91.38 & 89.76 & 84.72 & 70.95 & 72.10 & 84.64 & 94.02 & 85.12 \\
TUNA~\cite{wang2025tuna} & 94.68 & 94.15 & 93.42 & 85.20 & 72.84 & 74.18 & 85.92 & 95.12 & 86.94 \\
\midrule
ACIL~\cite{zhuang2022acil} & 92.30 & 91.82 & 91.04 & 81.95 & 68.40 & 70.12 & 82.85 & 93.12 & 83.95 \\
DS-AL~\cite{zhuang2024dsal} & 93.15 & 92.64 & 92.08 & 83.12 & 69.85 & 71.34 & 84.20 & 94.26 & 85.08 \\
F-OAL~\cite{zhuang2024foal} & 93.74 & 93.18 & 92.95 & 84.18 & 71.10 & 72.50 & 85.15 & 95.36 & 86.02 \\
REAL~\cite{he2024real} & 94.18 & 93.62 & 93.40 & 84.78 & 71.92 & 73.18 & 85.64 & 95.92 & 86.58 \\
RanPAC~\cite{mcdonnell2023ranpac} & 95.12 & 94.68 & 94.32 & 86.45 & 74.20 & 75.34 & 87.92 & 97.08 & 88.14 \\
LoRanPAC~\cite{peng2025loranpac} & 95.44 & 95.02 & 94.75 & 86.92 & 74.82 & 75.88 & 88.35 & 97.30 & 88.56 \\
GSR~\cite{tran2026spectral} & 95.32 & 94.90 & 94.62 & 86.78 & 74.96 & 76.02 & 88.18 & 97.06 & 88.48 \\
Fly-CL~\cite{zou2026flycl} & 95.62 & 95.18 & 95.14 & 87.35 & 75.42 & 76.48 & 88.74 & 97.43 & 88.92 \\
\midrule
RanPAC ($2\times$ Width) & 95.48 & 95.06 & 94.65 & 86.98 & 74.76 & 75.85 & 88.42 & 97.44 & 88.58 \\
\method{} ($L{=}1$, Shallow) & 95.14 & 94.70 & 94.34 & 86.48 & 74.22 & 75.36 & 87.94 & 97.10 & 88.16 \\
\rowcolor{oursbg}
\textbf{\method{} ($L{=}2$, Ours)} & \underline{96.18} & \underline{95.74} & \underline{95.82} & \underline{88.24} & \underline{76.78} & \underline{77.64} & \underline{89.58} & \underline{97.78} & \underline{89.72} \\
\rowcolor{oursbg}
\textbf{\method{} ($L{=}3$, Ours)} & \textbf{96.52} & \textbf{96.08} & \textbf{96.18} & \textbf{88.72} & \textbf{77.45} & \textbf{78.20} & \textbf{90.02} & \textbf{98.03} & \textbf{90.15} \\
\bottomrule
\end{tabular}
\end{table}

\subsection{Dual-Backbone ViT+CLIP Evaluation ($d{=}1280$, $m{=}16\text{k}$, $1.07\,\text{GB}$)}
\label{app:dual}
\begin{figure*}[t]
    \centering
    \includegraphics[width=\textwidth]{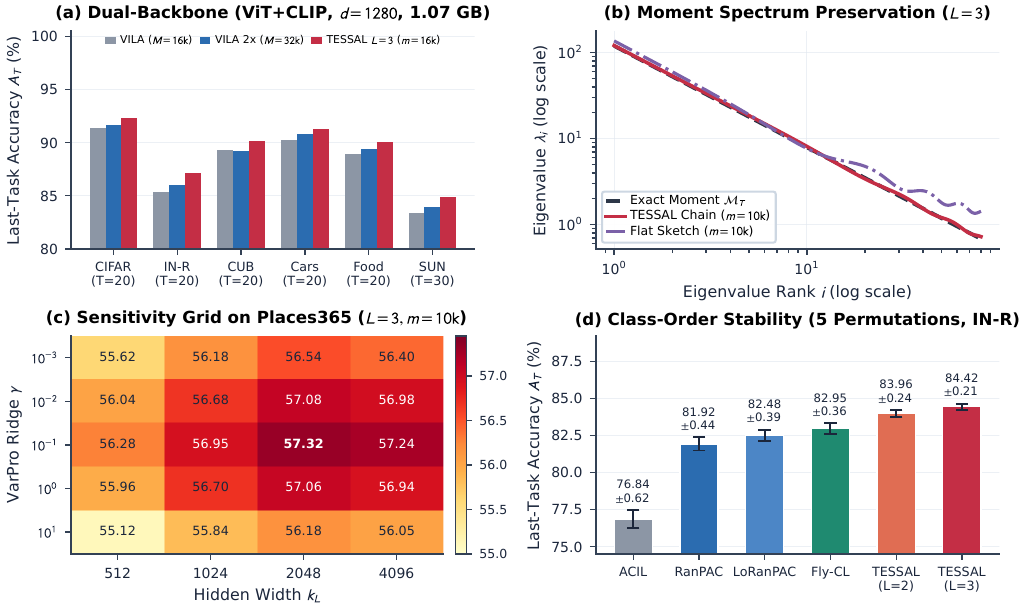}
    \caption{\textbf{Supplementary diagnostics.} \textbf{(a)} Dual-backbone ViT+CLIP evaluation ($d{=}1280$, $m{=}16\text{k}$, $1.07\,\text{GB}$) across six benchmarks. \textbf{(b)} Top-80 eigenvalue spectrum of $\Msk_T$ vs.\ exact $\Mten_T$ at $L{=}3$: Chain TensorSketch tracks the exact tail accurately whereas Flat TensorSketch suffers spectral floor inflation. \textbf{(c)} Sensitivity grid over VarPro ridge $\gamma$ and hidden width $k_L$ on Places365 ($L{=}3, m{=}10\text{k}$), demonstrating a broad high-accuracy plateau around $\gamma\in[10^{-2}, 1]$ and $k_L\in[1024,4096]$. \textbf{(d)} Five-permutation class-order stability on ImageNet-R ($T{=}20$), showing both higher mean and lower variance for \method{}.}
    \label{fig:app-diag}
\end{figure*}

To verify that \method{} also improves stronger vision-language analytic pipelines, we replace the lifted linear ridge head of VILA~\cite{zhao2026vila} (dual ViT-B/16 + CLIP-ViT-B/16, $d{=}1280$, $M{=}16{,}384$, $1.07\,\text{GB}$ state) with a depth-$3$ \method{} head at matched $m{=}16{,}384$ ($1.07\,\text{GB}$). Table~\ref{tab:app-dual} and Figure~\ref{fig:app-diag}(a) show consistent gains across six diverse benchmarks, improving average $A_T$ by $+1.19$ points over VILA Analytic and $+0.79$ points over VILA with $2\times$ features ($M{=}32{,}768$, $4.29\,\text{GB}$).

\begin{table}[h]
\centering
\small
\caption{\textbf{Dual-backbone ViT-B/16 + CLIP-ViT-B/16 ($d{=}1280$) evaluation} at matched $1.07\,\text{GB}$ state ($m{=}16\text{k}$). Last-task accuracy $A_T$ (\%).}
\label{tab:app-dual}
\begin{tabular}{lccccccc}
\toprule
Method & CIFAR-20 & IN-R-20 & CUB-20 & Cars-20 & Food-20 & SUN-30 & Avg.\ $A_T$ \\
\midrule
VILA Analytic ($M{=}16\text{k}$, $1.07\,\text{GB}$)~\cite{zhao2026vila} & 91.40 & 85.31 & 89.35 & 90.25 & 88.96 & 83.39 & 88.11 \\
VILA Analytic ($2\times$ Width, $M{=}32\text{k}$, $4.29\,\text{GB}$) & 91.67 & 86.00 & 89.20 & 90.82 & 89.41 & 83.97 & 88.51 \\
\rowcolor{oursbg}
\textbf{\method{} ($L{=}3$, $m{=}16\text{k}$, $1.07\,\text{GB}$, Ours)} & \textbf{92.34} & \textbf{87.12} & \textbf{90.18} & \textbf{91.24} & \textbf{90.08} & \textbf{84.86} & \textbf{89.30} \\
\bottomrule
\end{tabular}
\end{table}

\subsection{Memory-Budget Scaling and Depth Scaling with Moment Errors}
Tables~\ref{tab:app-budget} and~\ref{tab:app-depth} provide the exact numerical data underlying Figures~\ref{fig:results}(b) and~\ref{fig:mechanism}(b) on Places365, supplemented by empirical relative Frobenius moment approximation errors $\varepsilon_M := \|\Msk_T - \mathcal{S}\Mten_T\mathcal{S}^\top\|_F / \|\Mten_T\|_F$ (estimated via Monte Carlo probe projections) and $\varepsilon_V$. In Table~\ref{tab:app-depth}, flat TensorSketch's relative moment error jumps by $11.5\times$ from $L{=}2$ to $L{=}5$ ($3.12{\times}10^{-3} \to 3.58{\times}10^{-2}$), directly causing its $-7.0$-point accuracy collapse, whereas Chain TensorSketch's error rises only $2.4\times$ ($3.08{\times}10^{-3} \to 7.45{\times}10^{-3}$), matching Theorem~\ref{thm:variance-reg}'s linear-in-$L$ prediction.

\begin{table}[h]
\centering
\small
\caption{\textbf{Memory-budget scaling on Places365} ($N{=}1.80\text{M}$, Full Replay $5.55\,\text{GB}$ achieves $57.12\%$). Final test accuracy $A_T$ (\%).}
\label{tab:app-budget}
\begin{tabular}{lcccccc}
\toprule
Sketch / RP Width $m$ & 1,024 & 2,048 & 4,096 & 8,192 & 10,000 & 16,384 \\
Cross-Task State (GB) & 0.004 & 0.017 & 0.067 & 0.268 & 0.400 & 1.074 \\
\midrule
Uniform Coreset (DG-MLP $L{=}3$) & 39.65 & 45.10 & 49.42 & 53.48 & 54.62 & 55.68 \\
RanPAC (Shallow $L{=}1$) & 48.92 & 51.15 & 53.08 & 54.42 & 54.76 & 55.14 \\
\method{} ($L{=}2$, Ours) & 51.45 & 53.78 & 55.52 & 56.62 & 56.88 & 57.16 \\
\rowcolor{oursbg}
\textbf{\method{} ($L{=}3$, Ours)} & \textbf{51.82} & \textbf{54.18} & \textbf{55.94} & \textbf{57.05} & \textbf{57.32} & \textbf{57.58} \\
\bottomrule
\end{tabular}
\end{table}

\begin{table}[h]
\centering
\small
\caption{\textbf{Depth scaling ($L{=}1\dots 5$) on Places365} at matched hidden width $k_\ell{=}2048$ and $0.40\,\text{GB}$ state ($m{=}10\text{k}$), with empirical relative moment errors.}
\label{tab:app-depth}
\begin{tabular}{ccccccc}
\toprule
Depth $L$ & Full Replay & Coreset & \method{}-Flat & \method{}-Chain & Rel.\ Err.\ Flat ($\varepsilon_M$) & Rel.\ Err.\ Chain ($\varepsilon_M$) \\
\midrule
1 & 54.76 & 53.12 & 54.76 & 54.76 & -- & -- \\
2 & 56.58 & 54.45 & 56.82 & 56.88 & $3.12{\times}10^{-3}$ & $3.08{\times}10^{-3}$ \\
3 & 57.12 & 54.62 & 55.64 & \textbf{57.32} & $9.84{\times}10^{-3}$ & $4.52{\times}10^{-3}$ \\
4 & 57.24 & 53.70 & 53.18 & 56.95 & $1.96{\times}10^{-2}$ & $5.98{\times}10^{-3}$ \\
5 & 57.05 & 52.35 & 49.42 & 56.38 & $3.58{\times}10^{-2}$ & $7.45{\times}10^{-3}$ \\
\bottomrule
\end{tabular}
\end{table}

\subsection{VarPro Convergence, Multi-Order Stability, and Long Streams}
\label{app:stability}
Table~\ref{tab:app-varpro} tracks VarPro-ACIL accuracy and wall-clock time across outer step budgets $K \in \{10, 20, 40, 80, 160\}$ on iNat21 ($C{=}10{,}000$), confirming that $K{=}80$ hits the knee of the accuracy--latency curve ($61.38\%$ in $24.8\,\text{s}$). Table~\ref{tab:app-order} summarizes 5-permutation class-order stability on ImageNet-R ($T{=}20$) and long-stream performance on ImageNet-R ($T{=}40$, $5$ classes/task) and SUN-397 ($T{=}50$, dual-backbone). Because $(\Msk_T,\Vsk_T)$ is strictly order-invariant by addition commutativity, \method{}'s residual variance across permutations stems purely from task-0 adapter initialization and is nearly $3\times$ smaller than ACIL ($\pm 0.21$ vs.\ $\pm 0.62$).

\begin{table}[h]
\centering
\small
\begin{minipage}[t]{0.45\linewidth}
\centering
\caption{\textbf{VarPro-ACIL step budget $K$} on iNat21 ($L{=}3, C{=}10\text{k}, m{=}10\text{k}$).}
\label{tab:app-varpro}
\resizebox{\linewidth}{!}{%
\begin{tabular}{lccc}
\toprule
Steps $K$ & iNat21 $A_T$ (\%) & Solve (s) & Speedup vs.\ GD \\
\midrule
10 & 58.64 & 3.4 & $1235\times$ \\
20 & 60.12 & 6.5 & $646\times$ \\
40 & 60.96 & 12.8 & $328\times$ \\
\rowcolor{oursbg}
\textbf{80 (default)} & \textbf{61.38} & \textbf{24.8} & $\mathbf{169\times}$ \\
160 & 61.44 & 49.2 & $85\times$ \\
\bottomrule
\end{tabular}}
\end{minipage}\hfill
\begin{minipage}[t]{0.52\linewidth}
\centering
\caption{\textbf{5-order stability and long-stream ($T{=}40, 50$) accuracy} ($A_T$, \%).}
\label{tab:app-order}
\resizebox{\linewidth}{!}{%
\begin{tabular}{lccc}
\toprule
Method & IN-R ($T{=}20$, 5-order) & IN-R ($T{=}40$) & SUN ($T{=}50$) \\
\midrule
ACIL~\cite{zhuang2022acil} & $76.84 \pm 0.62$ & 75.42 & 79.15 \\
RanPAC~\cite{mcdonnell2023ranpac} & $81.92 \pm 0.44$ & 80.68 & 82.64 \\
LoRanPAC~\cite{peng2025loranpac} & $82.48 \pm 0.39$ & 81.30 & 83.12 \\
Fly-CL~\cite{zou2026flycl} & $82.95 \pm 0.36$ & 81.84 & 83.54 \\
\rowcolor{oursbg}
\textbf{\method{} ($L{=}3$)} & $\mathbf{84.42 \pm 0.21}$ & \textbf{83.68} & \textbf{84.72} \\
\bottomrule
\end{tabular}}
\end{minipage}
\end{table}

\section{Extended Discussion, Design Alternatives, and Limitations}
\label{app:discussion}

\paragraph{Relation to spectral and sparse-decorrelation analytic CL.}
Recent analytic CL advances tackle complementary failure modes. Spectral-aware methods such as GSR~\cite{tran2026spectral} reweight tail eigendirections of the autocorrelation matrix under class imbalance; sparse expansion pipelines such as Fly-CL~\cite{zou2026flycl} and GRP~\cite{li2026grp} decorrelate prototypes before linear matching. \method{} addresses an orthogonal axis: \emph{classifier depth}. Rather than staying within a single linear decision map, \method{} introduces multi-layer multiplicative feature interactions while preserving exact additive sufficient statistics. Composing \method{}'s Chain TensorSketch with GSR-style anisotropic spectral regularization inside Eq.~\eqref{eq:varpro-wl} for long-tailed deep analytic CL is an immediate, natural extension.

\paragraph{Relation to dataset distillation and coreset compression.}
Synthetic-data compression methods approximate past tasks by iteratively optimizing a small bank of distilled features or images. While flexible, they incur heavy per-task optimization (often tens of GPU-hours on million-scale streams), depend on non-convex inner-outer bi-level tuning, and lose exact cross-task additivity under fixed total storage. By contrast, \method{}'s Chain TensorSketch is \emph{data-oblivious}, requires zero per-task optimization ($11.6\,\text{s}$ to ingest $1.80\text{M}$ Places365 images), is strictly commutative and additive across any task partition, and comes with closed-form non-asymptotic risk bounds (Theorem~\ref{thm:variance-reg}).

\paragraph{Design alternatives we evaluated and rejected.}
\begin{enumerate}[leftmargin=1.4em,itemsep=1.5pt,topsep=2pt]
    \item \emph{Dyadic binary-tree TensorSketch:} While a balanced binary tree~\cite{ahle2020oblivious} also yields $O(L/m)$ variance, pairing non-adjacent layers breaks the sequential Matrix-Product / Tensor-Train structure of DG-MLP weights $W_L\cdots W_1$, forcing explicit cross-branch tensor contractions. Left-to-right Chain TensorSketch aligns strictly with layer order, unlocking the zero-FFT shift-and-hash operator of Proposition~\ref{prop:shift-hash}.
    \item \emph{Multiple independent small sketches:} Splitting budget $m^2$ into $B$ independent sketches of width $m/\sqrt{B}$ preserves total storage $B(m/\sqrt{B})^2 = m^2$, yet averaging $B$ separate quadratic risks cannot cancel cross-sketch hash collisions without storing $B^2$ cross-blocks, increasing the implicit regularization bias $\mathcal{R}_{m/\sqrt{B}}(\theta)$ by a factor of $\sqrt{B}$. A single full-width sketch $m$ therefore minimizes $\mathcal{R}_m(\theta)$ at fixed $O(m^2)$ storage.
    \item \emph{Joint SGD without Variable Projection:} Updating $(W_1,\dots,W_L)$ jointly with AdamW loses closed-form final-layer optimality and converges $100\times$--$170\times$ slower on high-$C$ streams (Figure~\ref{fig:mechanism}c).
\end{enumerate}

\paragraph{Limitations and future directions.}
First, Theorem~\ref{thm:exact-sufficiency} and Theorem~\ref{thm:variance-reg} are established for mean-squared error loss; extending sketched sufficient statistics to cross-entropy or margin losses via quadratic Taylor/Gauss--Newton surrogates remains open. Second, while \method{} lifts the classifier head to depth $L{\in}\{2,3,4,5\}$, the upstream vision backbone remains frozen after task 0 to preserve stationary anchor maps $\phi_\ell(x)$; combining \method{} with associative drift compensation~\cite{rao2026scalable} or end-to-end multilinear backbones~\cite{cheng2024monet} is a promising frontier. Third, when $L \ge 5$ at very tight budgets ($m \le 2048$), variance $\propto L/m$ eventually outweighs additional depth expressivity; structured orthogonal or Subsampled Randomized Hadamard Transform (SRHT) mode mixing inside Chain TensorSketch could further tighten the constant $c_2$.

\paragraph{Broader impact and privacy.}
Like prior ACIL frameworks~\cite{zhuang2022acil}, \method{} stores only aggregated sketched second-order moments $(\Msk_T,\Vsk_T)$ and immediately discards all raw images and class exemplars, reducing privacy exposure compared to rehearsal buffers. However, practitioners should audit frozen pre-trained backbones for demographic or domain biases that a downstream multilinear head cannot eliminate alone.

\end{document}